\documentclass[10pt,twocolumn]{article}

\usepackage{algorithm, algorithmic}
\usepackage{amsmath, amssymb, mathtools}
\usepackage{graphicx}
\usepackage{multicol}
\usepackage{placeins}
\usepackage[caption=false]{subfig}
\usepackage{url}
\usepackage{xfrac}
\usepackage{xcolor}
\usepackage{booktabs}
\usepackage{authblk}
\usepackage[margin=0.6in]{geometry}

\newcommand{\ra}[1]{\renewcommand{\arraystretch}{#1}}
\newcommand{\ca}[1]{\setlength{\tabcolsep}{#1pt}}
\begin{document}

\title{Auto-adaptative Laplacian Pyramids for High-dimensional Data Analysis}

\author[1]{\'{A}ngela Fern\'{a}ndez\thanks{a.fernandez@uam.es}}
\author[2]{Neta Rabin\thanks{netar@afeka.ac.il}}
\author[2]{Dalia Fishelov\thanks{daliaf@post.tau.ac.il}}
\author[1]{Jos\'{e} R. Dorronsoro\thanks{jose.dorronsoro@uam.es}}
\affil[1]{Dpto. Ing. Inform\'{a}tica, UAM, 28049, Madrid, Spain}
\affil[2]{Dept. Exact Sciences, Afeka, Tel-Aviv, 69107, Israel}

\renewcommand\Authands{ and }
\date{}

\maketitle

\begin{abstract}
Non-linear dimensionality reduction techniques such as manifold learning algorithms have become a common way for processing and analyzing high-dimensional patterns that often have attached a target that corresponds to the value of an unknown function. Their application to new points consists in two steps: first, embedding the new data point into the low dimensional space and then, estimating the function value on the test point from its neighbors in the embedded space.

However, finding the low dimension representation of a test point, while easy for simple but often not powerful enough procedures such as PCA, can be much more complicated for methods that rely on some kind of eigenanalysis, such as Spectral Clustering (SC) or Diffusion Maps (DM). 
Similarly, when a target function is to be evaluated, averaging methods like nearest neighbors may give unstable results if the function is noisy.
Thus, the smoothing of the target function with respect to the intrinsic, low-dimensional representation that describes the geometric structure of the examined data is a challenging task.

In this paper we propose Auto-adaptive Laplacian Pyramids (ALP), an extension of the standard Laplacian Pyramids model that incorporates a modified Leave One Out cross validation (LOOCV) procedure that avoids the large cost of standard LOOCV and offers the following advantages: (i) it selects automatically the optimal function resolution (stopping time) adapted to the data and its noise, (ii) it is easy to apply as it does not require parameterization, (iii) it does not overfit the training set and (iv) it adds no extra cost compared to other classical interpolation methods. We illustrate numerically ALP's behavior on a synthetic problem and apply it to the computation of the DM projection of new patterns and to the extension to them of target function values on a radiation forecasting problem over very high dimensional patterns.

{\bf Keywords:} Laplacian Pyramids, LOOCV, Manifold Learning, Diffusion Maps.
\end{abstract}

\section{Motivation}
\label{sec:intro}
An important challenge in data mining and machine learning is the proper analysis of a given dataset, especially for understanding and working with functions defined over it.
In particular, manifold learning algorithms have become a common way for processing and analyzing high-dimensional data and the so called ``diffusion analysis'' allows us to find the most appropriate geometry to study such functions~\cite{SzlamRegularized}.
These methods are based on the construction of a diffusion operator that depends on the local geometry of the data, which is then used to embed the high-dimensional points into a lower-dimensional space maintaining their geometric properties and, hopefully, making easier the analysis of functions over it.
On the other hand, extending functions in such an embedding for new data points may be challenging, either because of the noise or the presence of low-density areas that make insufficient the number of available training points. Also it is difficult to set the neighborhood size for new, unseen points as it has to be done according to the local behavior of the function.

The classical methods for function extension like Geometric Harmonics~\cite{GHCoifman} have parameters that need to be carefully set, and in addition there does not exist a robust method of picking the correct neighborhood in the embedding for function smoothing and evaluation.
A first attempt to simplify these approaches was Laplacian Pyramids (LP), a multi-scale model that generates a smoothed version of a function in an iterative manner by using Gaussian kernels of decreasing widths~\cite{LPRabin}. It is a simple method for learning functions from a general set of coordinates and can be also applied to extend embedding coordinates, one of the big challenges in diffusion methods. 
Recently \cite{NLICAAizenbud} introduced a geometric PCA based out-of-sample extension for the purpose of adding new points to a set of constructed embedding coordinates. 

A na\"ive way to extend the target function to a new data point could be to find the point's nearest neighbors (NN) in the embedded space and average their function values. The NN method for data lifting was compared in~\cite{LPDsilva} with the LP version that was proposed in ~\cite{LPRabin}, and this last method performed better than NN. Buchman et al.~\cite{Buchman11} also described a different, point-wise adaptive approach, which requires setting the nearest neighborhood radius parameter for every point.

Nevertheless, and as it is often the case in machine learning, when we apply the previous LP model, we can overfit the data if we try to refine too much the prediction during the training phase.
In fact, it is difficult to decide when to stop training to obtain good generalization capabilities.
A usual approach is to apply the Cross Validation (CV)~\cite[chap. 9]{Duda} method to get a validation error and to stop when this error starts to increase. 
An extreme form of CV is the Leave One Out CV (LOOCV): a model is built using all the samples but one, which is then used as a single validation pattern; this is repeated for each sample in the dataset, and the validation error is the average of all the errors.
Although LOOCV has a theoretical backing and often yields good results, it has the drawback of a big computational cost, though not in some important cases (see Section \ref{sec:AALP}).

\begin{figure}[ht]
\centering
	\includegraphics[width=0.45\textwidth]{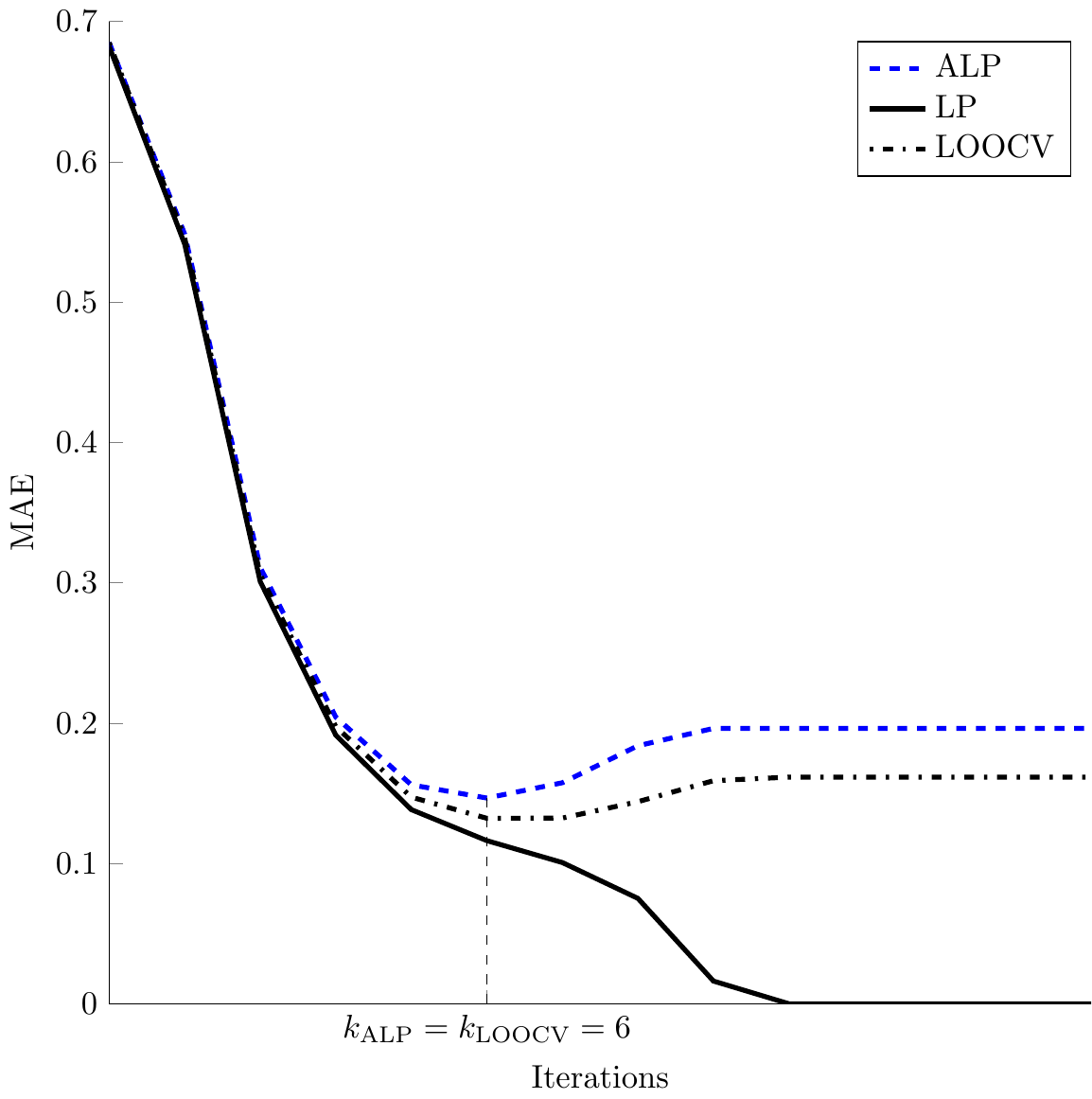}
\caption{Training and LOOCV errors for the original and modified LP models applied to a synthetic example consisting on a perturbed sine shown in Section~\ref{sec:Synt}.} 
\label{motivation}
\end{figure}

In this paper we propose Auto-adaptive LP (ALP), a modification in the LP training algorithm that merges training and approximate LOOCV in one single phase. To do so we simply build the kernel matrix with zeros in its diagonal. As we shall see, with this change we can implement an LOOCV approximation without any additional cost during the training step.
This reduces significantly training complexity and provides an automatic criterion to stop training so that we greatly avoid the risk of severe overfitting that may appear in standard LP.
This effect can be observed in Figure~\ref{motivation} when our LP proposal is applied to the synthetic example used in Section~\ref{sec:Synt}. The solid and dashed black lines represent the LP training error and the LOOCV error per iteration respectively, and the dashed blue line represents the error for our proposed method. 
The blue line, that corresponds to the ALP training error attains its minimum at the same iteration prescribed by LOOCV  for LP.

Therefore, ALP doesn't overfit the data and, moreover, doesn't essentially require any parametrization or expert knowledge about the problem, while still achieving a good test error. Moreover, it adds no extra cost compared to other classical neighbor-based interpolation methods.

This paper is organized as follows. In Section~\ref{sec:LP} we briefly review the LP model and present a detailed  analysis of its training error. We describe ALP in Section~\ref{sec:AALP}, which we apply to a synthetic example in Section~\ref{sec:Synt} and to a real world example in Section~\ref{sec:Real}. The paper ends with some conclusions in Section~\ref{sec:Concl}.

\section{The Laplacian Pyramids}
\label{sec:LP}
The Laplacian Pyramid (LP) is an iterative model that was introduced by Burt and Adelson \cite{LPBurt} for its application in image processing and, in particular, image encoding. In the traditional algorithm, LP decomposes the input image into a series of images, each of them capturing a different frequency band of the original one. This process is carried out by constructing Gaussian-based smoothing masks of different widths, followed by a down-sampling (quantization) step. LP was later proved to be a tight frame (see Do and Vetterly \cite{LPDo}) and used for signal processing applications, for example as a reconstruction scheme in~\cite{LiftingLP}.

In \cite{LPRabin}, it was introduced a multi-scale algorithm in the spirit of LP to be applied in the setting of high-dimensional data analysis. In particular, it was proposed as a simple method for extending low-dimensional embedding coordinates that result from the application of a non-linear dimensionality reduction technique to a high-dimensional dataset (recently applied in \cite{Mishne13}). 
   
We review next the LP procedure as described in \cite{LPRabin} (the down-sampling step, which is part of Burt and Adelson's algorithm is skipped here).
Let $S=\{x_i\}_{i=1}^{N} \in \mathbb{R}^N$ be the sample dataset; the algorithm approximates a function $f$ defined over $S$ by constructing a series of functions $\{\tilde{f}_i\}$ obtained by several refinements $d_i$ over the error approximations. 
In a slight abuse of language we will use the same notation $f$ for both the general function $f(x)$ and also for the vector of its sample values $f =(f_1 = f(x_1), \ldots, f_N = f(x_N))$.
The result of this process gives a function approximation
\begin{equation*}
	f \simeq \tilde{f} = \tilde{f}_0 + d_1 + d_2 + d_3 + \cdots 
\end{equation*}
In more detail, a first level kernel $K^0_\sigma(x,x') = \Phi \left( \text{dist}(x,x')/\sigma \right)$ is chosen using a wide, initial scale $\sigma$ and where $\text{dist}(x, x')$ denotes some distance function between points in the original dataset.
A usual choice and the one we will use here is the Gaussian kernel with Euclidean distances, i.e., to take $\text{dist}(x, x')=\|x - x'\| $ and then
\begin{equation*}
K^{0}(x, x') = \kappa e^{-\frac{\| x - x' \|^2}{\sigma^2}},
\end{equation*}
where $\kappa$ is the Gaussian kernel normalizing constant. 
As before, we will use the $K^0$ notation for both the general continuous kernel 
$K^{0}(x, x')$ and for its discrete matrix counterpart $K_{jk}^{(0)} = K^{0} (x_j, x_k)$ over the sample points.

The smoothing operator $P^{0}$ is constructed as the normalized row-stochastic matrix of the kernel
\begin{equation}
P^{0}_{ij} = \frac{K^0_{ij}}{\sum_k K^0_{ik}}.
\label{k2p}
\end{equation}
A first coarse representation of $f$ is then generated by the convolution $\tilde{f}_0 = f*P^{0}$ that captures the low-frequencies of the function. 
For the next steps we fix a $\mu > 1$, construct at step $i$ a sharper Gaussian kernel $P^i$ with scale $\sigma/\mu^i$, and the residual 
\begin{equation*}
d_{i-1} = f - \tilde{f}_{i-1},
\end{equation*}
which captures the error of the approximation to $f$ at the previous $i-1$ step, is used to generate a more detailed representation of $f$ given by
\begin{equation*}
\tilde{f}_i = \tilde{f}_{i-1} + d_{i-1}*P^{i} = \tilde{f}_{i-1} + g_{i-1},
\end{equation*}
with $g_{\ell} = d_{\ell} * P^{\ell+1}$.
The iterative algorithm stops once the norm of $d_i$ residual vector is smaller than a predefined error. 
Stopping at iteration $L$, the final LP model has thus the form
\begin{equation}
\tilde{f}_L = \tilde{f}_0 + \sum_0^{L-1} g_{\ell} = f*P^0 + \sum_0^{L-1} d_{\ell}*P^{\ell+1},
\label{LPmodel}
\end{equation} 
and extending this multi-scale representation to a new data point $x \in \mathbb{R}^N$ is now straightforward because we simply set 
\begin{align*}
\tilde{f}_L(x) & = f*P^0 (x) + \sum_0^{L-1} d_{\ell} * P^{\ell+1} (x) \\
& = \sum_j f_j P^0 (x, x_j) + \sum_1^{L} \sum_j d_{\ell-1; j} P^{\ell} (x, x_j).
\end{align*}
where we extend the $P^{\ell}$ kernels to a new $x$ as 
\begin{equation*}
P^{\ell} (x, x_j) = \frac{K^{\ell} (x, x_j)} {\sum_k K^{\ell} (x, x_k)}.
\end{equation*}
The overall cost is easy to analyze. Computing the convolutions $\tilde{f}_0 = f*P^0$, $g_{\ell} = d_{\ell}*P^{\ell-1}$ has a $O (N^2)$ cost for a size $N$ sample, while that of obtaining the $d_{\ell}$ is just $O(N)$. Thus, the overall cost of $L$ LP steps is $O(L N^2)$.

We observe that if we set a very small error threshold and run afterwards enough iterations, we will end up having $\tilde{f}_{\ell} = f$ over the training sample.
In fact, $\tilde{f}_{\ell} = \tilde{f}_{\ell-1} + g_{\ell-1}$ and, therefore, 
\begin{align*}
\tilde{f}_{\ell} & = \tilde{f}_{\ell-1} + g_{\ell-1} = \tilde{f}_{\ell-1} + (f - \tilde{f}_{\ell-1})*P^{\ell} \\
& = f*P^{\ell} + \tilde{f}_{\ell-1}*(I - P^{\ell}),
\end{align*}
with $I$ denoting the identity matrix.
Now, if we have $\tilde{f}^{\ell} \rightarrow \phi$, it follows taking limits that
\begin{equation*}
\phi = f* \lim P^{\ell} + \phi * \lim (I - P^{\ell})
\end{equation*}
i.e., $\phi = f$, for $P^{\ell} \rightarrow I$.

In practice, we will numerically have $P^{\ell} = I$ as soon as $\ell$ is large enough so that we have
$K^{\ell} (x_i, x_j) \simeq 0$.
We then have $d_{\ell; j} = 0$ for all $j$ and the LP model doesn't change anymore.
In other words, care has to be taken when deciding to stop the LP iterations to avoid overfitting.
In fact, we show next that when using Gaussian kernels, as we do, the $L_2$ norm of the LP errors  $\hat{d}_{\ell}$ decay extremely fast.

First notice that working in the continuous kernel setting, we have $P= K$ for a Gaussian kernel, for
then the denominator in \eqref{k2p} is just $\int K(x,z) dz = 1$.
Assume that $f$ is in $L_2$, so $\int_{x}{f^{2}(x) \; dx} < \infty $.
The LP scheme is a relaxation process for which in the first step the function $f$ is approximated by
$\mathcal{G}^{0}(f)=f \ast P^0 \left(x\right)$. 
For all $\ell$, $P^\ell \left(x\right)$
is an approximation to a delta function satisfying
\begin{align}\label{eq_Err_app1}
\int{P^\ell \left(x\right)dx}&=1, \nonumber\\ 
\int{x P^\ell \left(x\right)dx}&=0, \\ 
\int{ x^2 P^\ell \left(x\right) dx} &\le 2C. \nonumber
\end{align}
In the second step $f$ is approximated by $\mathcal{G}^{0}(f)+\mathcal{G}^{1}(d_0)$, where $d_0 = \mathcal{G}^{0}(f)-f$ and $\mathcal{G}^{1}(d_0) = d_0 \ast P^1 \left(x\right)$, and so on.
Taking the Fourier transform of $P^\ell \left(x\right)$, we have (see~\cite{Fishelov})
\begin{equation}\label{eq_Err_app2}
\left| \hat{P}^\ell \left(\omega\right)\right| \
\leq 1 + \frac{\sigma^2}{2} \int {x^2} P^\ell \left(x\right) dx
\le 1+C\sigma^2,
\end{equation}
where we have used \eqref{eq_Err_app1}.

We first analyze the error $d_0(x)$ in the first step, which is defined by $d_0(x)=f\ast P^0 \left(x\right)-f$. Taking the Fourier transform of $d_0(x)$ and using \eqref{eq_Err_app2} we have
\begin{equation}\label{eq_Err_app3}
\left| \hat{d_0}(\omega) \right|= \left| \hat{f}(w) \right| \left| \hat{P}^0 (\omega)-1\right| \le C\sigma_0^2\left|\hat{f}(\omega)\right|.
\end{equation}
The error in the second step is
\begin{equation}\label{eq_Err_app4}
d_1(x)=d_0-\mathcal{G}_1(d_0)=\left(f\ast P^0 -f\right)-d_0\ast P^1.
\end{equation}
Taking the Fourier transform of \eqref{eq_Err_app4} yields
\begin{equation*}\label{eq_Err_app5}
\begin{split}
\left| \hat{d}_1(\omega) \right| = \left| \hat{d}_0(\omega)-\hat{d}_0(\omega)\hat{P}^1 (\omega) \right|\\= \left|\hat{d_0}(\omega)\right|\left| \hat{P}^1 (\omega) -1 \right|.
\end{split}
\end{equation*}
Using \eqref{eq_Err_app2} and \eqref{eq_Err_app3} we obtain
\begin{equation*}\label{eq_Err_app6}
\left| \hat{d}_1(\omega) \right| \le C \left|\hat{d}_0(\omega)\right|\sigma_1^2 \le C\sigma_0^2\sigma_1^2\left|\hat{f}(\omega)\right|.
\end{equation*}
Since $\sigma_1 = \frac{\sigma_0}{\mu}$ with $\mu > 1$, then $\left| \hat{d}_1(\omega)\right| \le C \sigma_0^2 \frac{\sigma_0^2}{\mu^2}\left|\hat{f}(\omega)\right|.$ 
Similarly, for the $\ell^{th}$ step the error is bounded by
\begin{equation*}\label{eq_Err_app7}
\left| \hat{d}_{\ell}(\omega) \right| \le C\sigma_0^2\left(\frac{\sigma_0^2}{\mu^2}\right)^{\ell}\left|\hat{f}(\omega)\right|.
\end{equation*}
By Parseval's equality we obtain
\begin{equation*}\label{eq_Err_app8}
\left\| d_{\ell}\right\|_{L^2} \le C\sigma_0^2\left(\frac{\sigma_0^2}{\mu^2}\right)^{\ell}\left\|f \right\|_{L^2}.
\end{equation*}
Thus, the error's $L_2$ decays faster than any algebraic rate. 

We see next how we can estimate a final iteration value $L$ that prevents overfitting without incurring on additional costs.

\section{Auto-adaptative Laplacian Pyramids}
\label{sec:AALP}

The standard way to prevent overfitting is to use an independent validation subset and to stop the above $\ell$ iterations as soon as the validation error on that subset starts to increase.
This can be problematic for small samples and introduces a random dependence on the choice of the particular validation subset; $k$-fold cross validation is then usually the standard choice, in which we randomly distribute the sample in $k$ subsets, and iteratively use $k-1$ subsets for training and the remaining one for validation.
In the extreme case when $k = N$, i.e., we use just one pattern for validation, we arrive at Leave One Out Cross Validation (LOOCV) and stop the training iterations when the LOOCV error starts to increase.
Besides its simplicity, LOOCV has the attractive of being an almost unbiased estimator of the true generalization error (see for instance \cite{LOOCVCawley, LOOCVElisseeff}), although with possibly a high variance \cite{LOOCVKohavi}.
In our case LOOCV can be easily applied using for training a $N \times N$ normalized kernel matrix $P_{(p)}$ which is just the previous matrix $K$ where we set to $0$ the $p$-th rows and columns when $x_p$ is held out of the training sample and used for validation.
The most obvious drawback of LOOCV is its rather high cost, which in our case is $N \times O(L N^2) = O(L N^3)$ cost. 
However, it is often the case for other models that there are ways to estimate the LOOCV error with a smaller cost.
This can be done exactly in the case of $k$-Nearest Neighbors \cite{LOOCVFukunaga} or of Ordinary Least Squares (\cite{OLSHastie}, Chapter 7), or approximately for Support Vector Machines \cite{SVMChapelle} or Gaussian Processes \cite{GPRasmussen}.

In order to alleviate it, notice first that when we removed $x_p$ from the training sample, the test value at $x_p$ of the $f^{(p)}$ extension built is
\begin{align*}
f^{(p)}_L (x_p) & = \sum_{j \neq p} f_j P^{0} (x_p, x_j) + \sum_{\ell=1}^{L} \sum_{j \neq p} d^{(p)}_{\ell-1; j} P^{\ell} (x_p, x_j) \\
& = \sum_j f_j \tilde{P}^0 (x_p, x_j) + \sum_{\ell=1}^{L} \sum_j {d}^{(p)}_{\ell-1; j} \tilde{P}^{\ell} (x_p, x_j),
\end{align*}
where $d^{(p)}_\ell$ are the different errors computed using the $P_{(p)}^{\ell}$ matrices and where $\tilde{P}$ is now just the matrix $P$ with its diagonal elements set to $0$, i.e. $\tilde{P}_{i,i} = 0$, $\tilde{P}_{i,j} = P_{i,j}$ when $j\neq i$.

This observation leads to the modification we propose on the standard LP algorithm given in \cite{LPRabin}, and which simply consist on applying the LP procedure described in Section \ref{sec:LP} but replacing the $P$ matrix by its 0-diagonal version $\tilde{P}$, computing then $\tilde{f}_0 = f*\tilde{P}^{0}$ at the beginning, and $\tilde{d}_{\ell} = f - \tilde{f}_{\ell}$, $\tilde{g}_{\ell} = \tilde{d}_{\ell} * \tilde{P}^{\ell+1}$ and $\tilde{f}_{\ell}$ vectors at each iteration.
We call this algorithm the Auto-adaptative Laplacian Pyramid (ALP). 

According to the previous formula for the $f^{(p)}_L (x_p)$, we can take the ALP values $\tilde{f}_{L,p} = \tilde{f}_{L} (x_p)$ given by 
\begin{equation*}
\tilde{f}_{L} (x_p)= \sum_j f_j \tilde{P}^0 (x_p, x_j) + \sum_{\ell=1}^{L} \sum_j \tilde{d}_{\ell-1; j} \tilde{P}^{\ell} (x_p, x_j),
\end{equation*}
as approximations to the LOOCV validation values $f^{(p)}_L (x_p)$.
But then we can approximate the square LOOCV error at iteration $L$ as 
\begin{equation*}
\sum_p (f(x_p) - f^{(p)}_L (x_p))^2 \simeq \sum_p (f(x_p) - \tilde{f}_{L,p} )^2 = \sum_p ( \tilde{d}_{L;p} )^2 ,
\end{equation*}
which is just the training error of ALP. 
In other words, working with the $\tilde{P}$ matrix instead of $P$,  the training error at step $L$ gives in fact an approximation to the LOOCV error at this step.
The cost of running $L$ steps of ALP is just $O(L N^2)$ and, thus, we gain the advantage of the exhaustive LOOCV without any additional cost on the overall algorithm.
The complete training procedure is presented in Algorithm~\ref{alg:train} and the test algorithm is shown in Algorithm~\ref{alg:test}. 
\begin{algorithm}[ht]
\caption{The ALP Training Algorithm}
\label{alg:train}
\begin{footnotesize} 
  \begin{algorithmic}[1]
  \REQUIRE $x_{\text{tr}}$, $y_{\text{tr}}$, $\sigma_0$, $\mu$.
  \ENSURE $(\{d_i\}, \sigma_0, \mu, k)$ {\color{blue} \% The trained model.}

  \STATE $\sigma \leftarrow \sigma_0$; $d_0 \leftarrow y_{\text{tr}}$.
  \STATE $\tilde{f}_0 \leftarrow 0$; $i \leftarrow 1$. 
  \WHILE{$ ({\text{err}_i} <{\text{err}_{i-1}} ) $}
	\STATE $K \leftarrow e^{-\sfrac{\|x_{\text{tr}}-x_{\text{tr}}\|^2}{\sigma^2}}$.
	\STATE $P_i \leftarrow$ normalize($K$).
	\STATE $\tilde{P} \leftarrow P$ with $0$-diagonal. {\color{blue} \% LOOCV.}
    \STATE $\tilde{f}_{i} \leftarrow \tilde{f}_{i-1} + d_{i-1} * \tilde{P}_i$.
	\STATE $d_i \leftarrow f - \tilde{f}_{i}$.    
    \STATE $\text{err}_i \leftarrow \sfrac{d_i}{s_{\text{tr}}}$, with $s_{\text{tr}}$ the number of patterns in $x_{\text{tr}}$.
	\STATE $\sigma \leftarrow \sfrac{\sigma}{\mu}$; $i \leftarrow i + 1$.
  \ENDWHILE 
  \STATE $k \leftarrow \min_{i} d_i$. {\color{blue} \% Optimal iteration.}
 \end{algorithmic}
\end{footnotesize}
\end{algorithm}
\begin{algorithm}[ht]
\caption{The ALP Testing Algorithm}
\label{alg:test}
\begin{footnotesize} 
  \begin{algorithmic}[1]
  \REQUIRE $x_{\text{tr}}$, $x_{\text{te}}$, $(\{d_i\}, \sigma_0, \mu, k)$.
  \ENSURE $\hat{y}_{\text{te}}$.
  
  \STATE $\hat{y}_{\text{te}} \leftarrow 0$; $\sigma \leftarrow \sigma_0$.
  \FOR{i=0 \TO k-1}
  	\STATE $K \leftarrow e^{-\sfrac{\|x_{\text{tr}}-x_{\text{te}}\|^2}{\sigma^2}}$.
  	\STATE $P_i \leftarrow$ normalize($K$).
  	\STATE $\hat{y}_{\text{te}} \leftarrow \hat{y}_{\text{te}} + d_i * P_i$.
  	\STATE $\sigma \leftarrow \sfrac{\sigma}{\mu}$.
  \ENDFOR
  \end{algorithmic}
\end{footnotesize}
\end{algorithm}

The obvious advantage of ALP is that when we evaluate the training error, we are actually estimating the LOOCV error after each LP iteration.
Therefore, the evolution of these LOOCV values tells us which is the optimal iteration to stop the algorithm, i.e., just when the training error approximation to the LOOCV error starts growing.
Thus, we do not only remove the danger of overfitting but can also use the training error as an approximation to the generalization error.
This effect can be seen in Figure~\ref{motivation} which illustrates the application of ALP to the synthetic problem described in the next section and where the optimum stopping time for ALP is exactly the same that the one that would give the LOOCV error and where training error stabilizes afterwards at a slightly larger value.

Moreover, ALP achieves an automatic selection of the width of the Gaussian kernel which makes this version of LP to be auto-adaptative as it does not require costly parameter selection procedures.
In fact, choosing as customarily done $\mu = 2$, the only required parameter would be the initial $\sigma$ but provided it is wide enough, its $\sigma/2^\ell$ scalings will yield an adequate final kernel width.

\begin{figure*}[ht!]
\centering
	\subfloat{\includegraphics[width=0.24\textwidth]{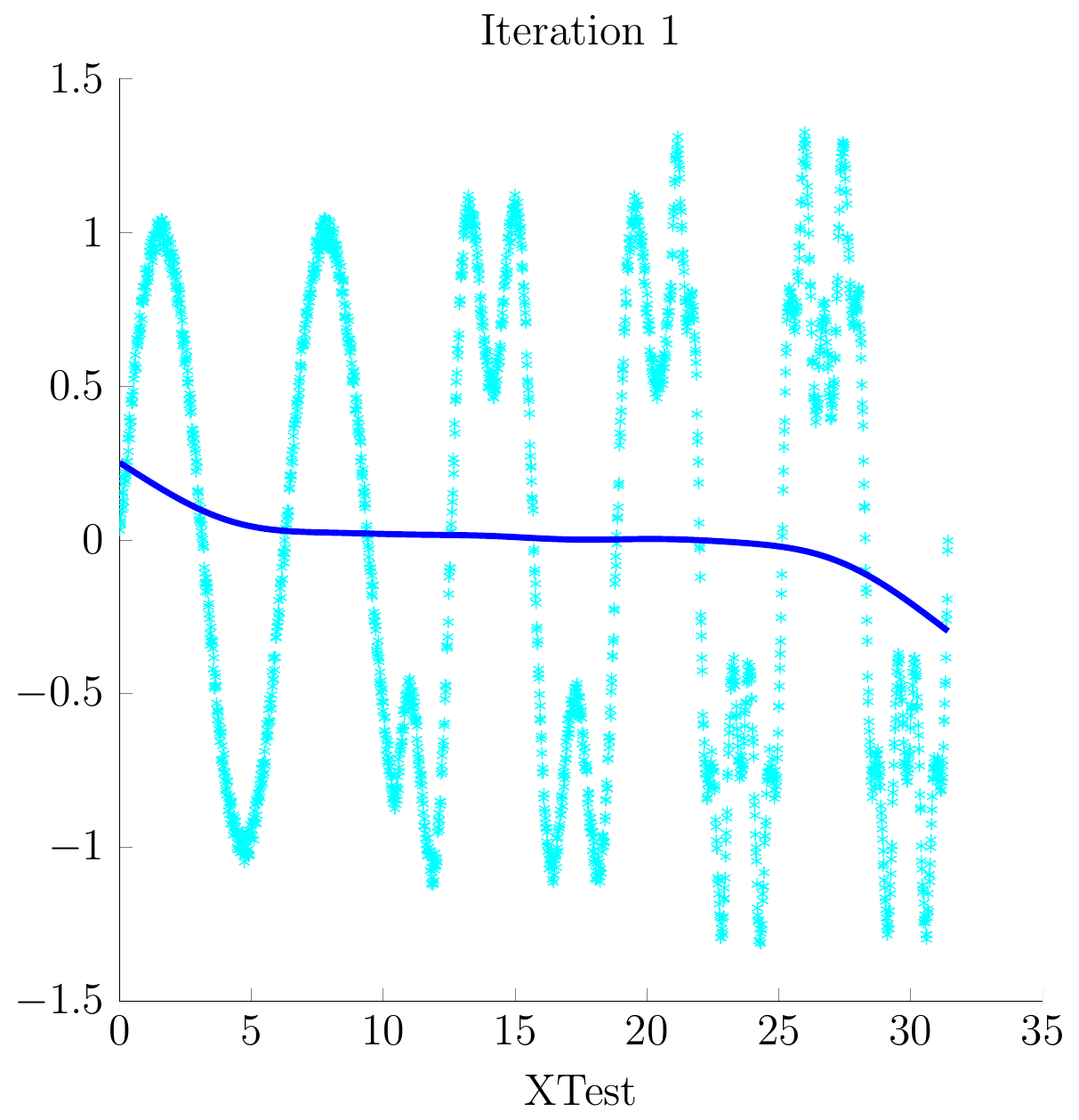}}
	\subfloat{\includegraphics[width=0.24\textwidth]{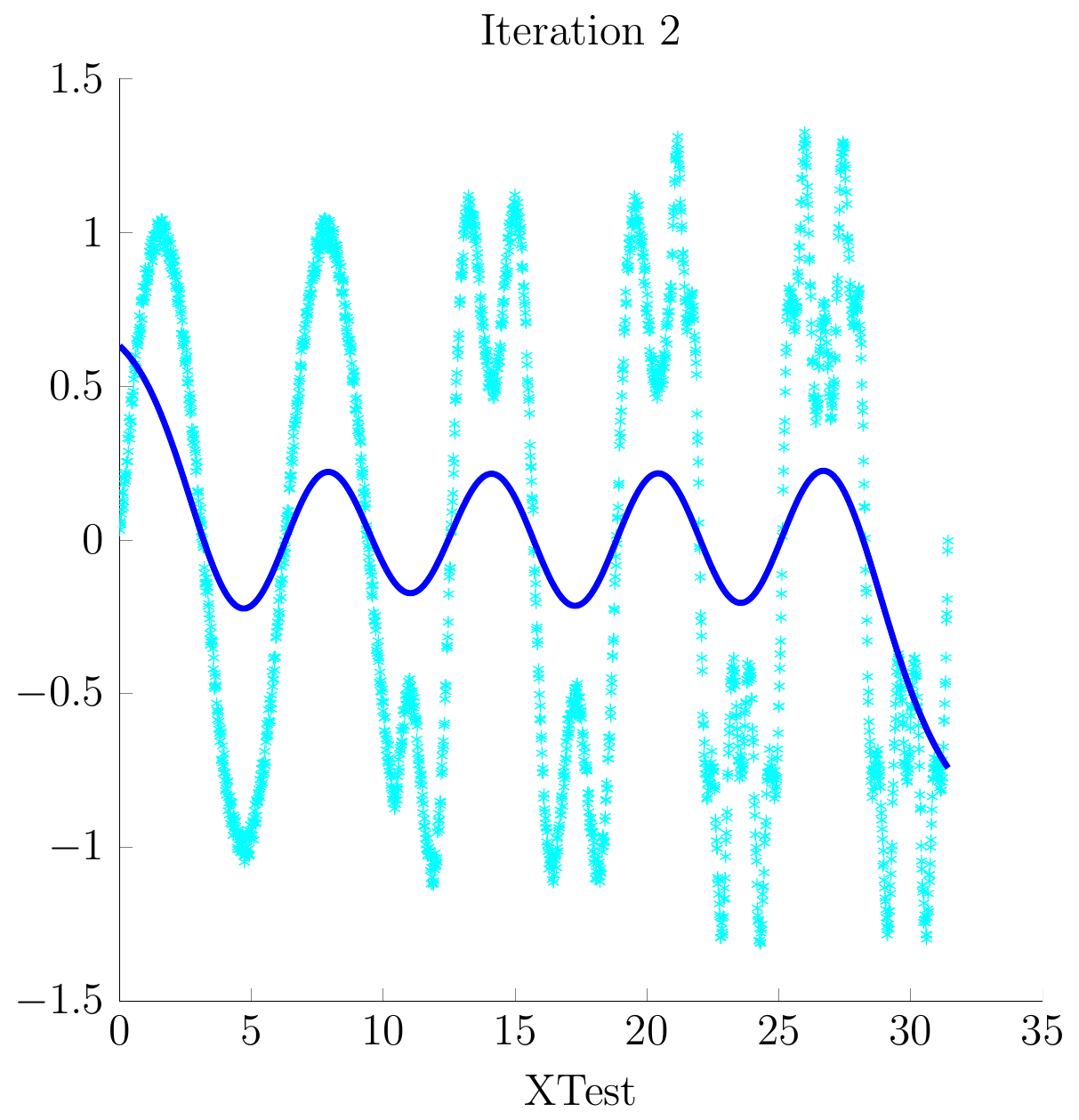}}
	\subfloat{\includegraphics[width=0.24\textwidth]{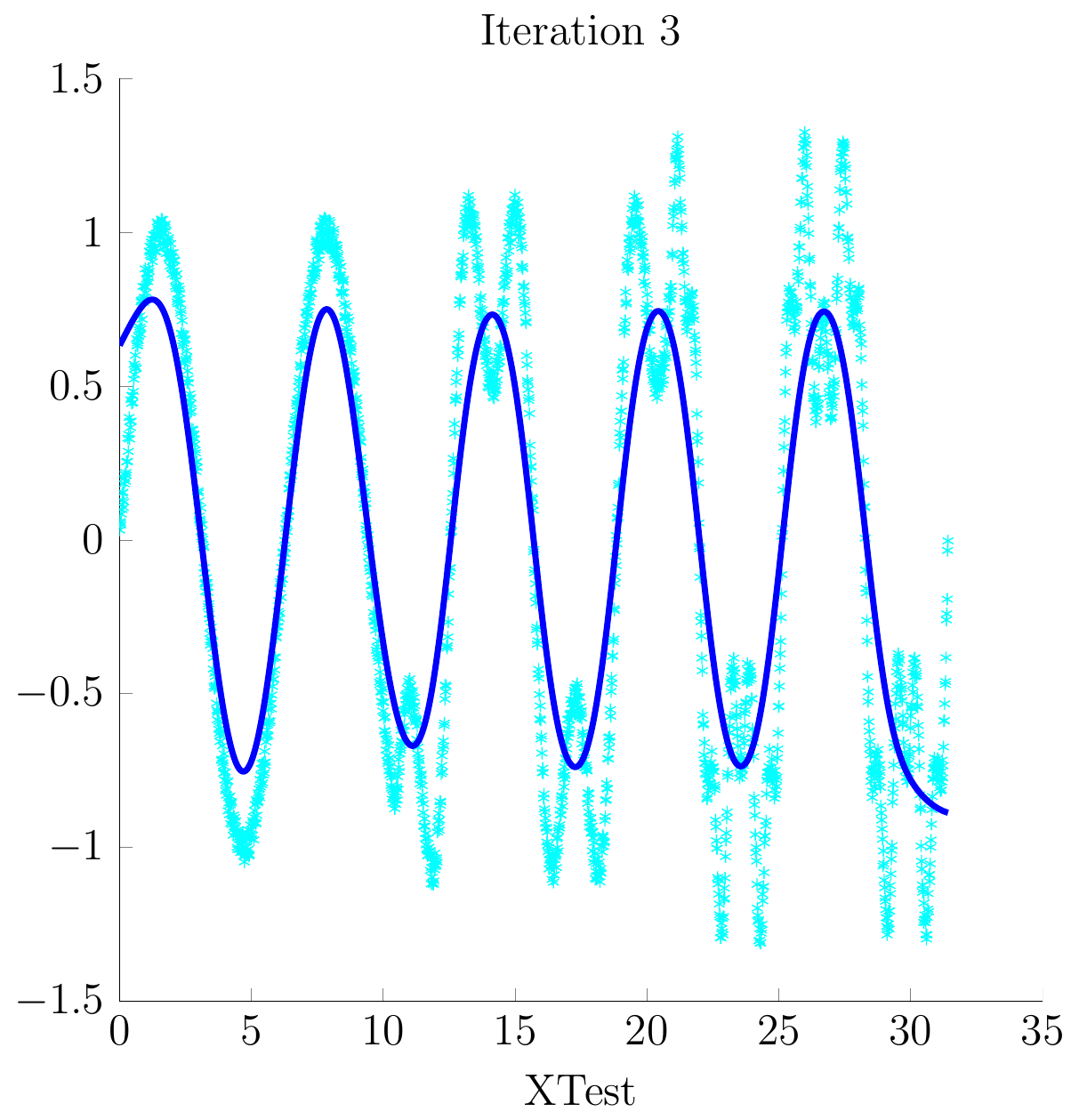}}
	\subfloat{\includegraphics[width=0.24\textwidth]{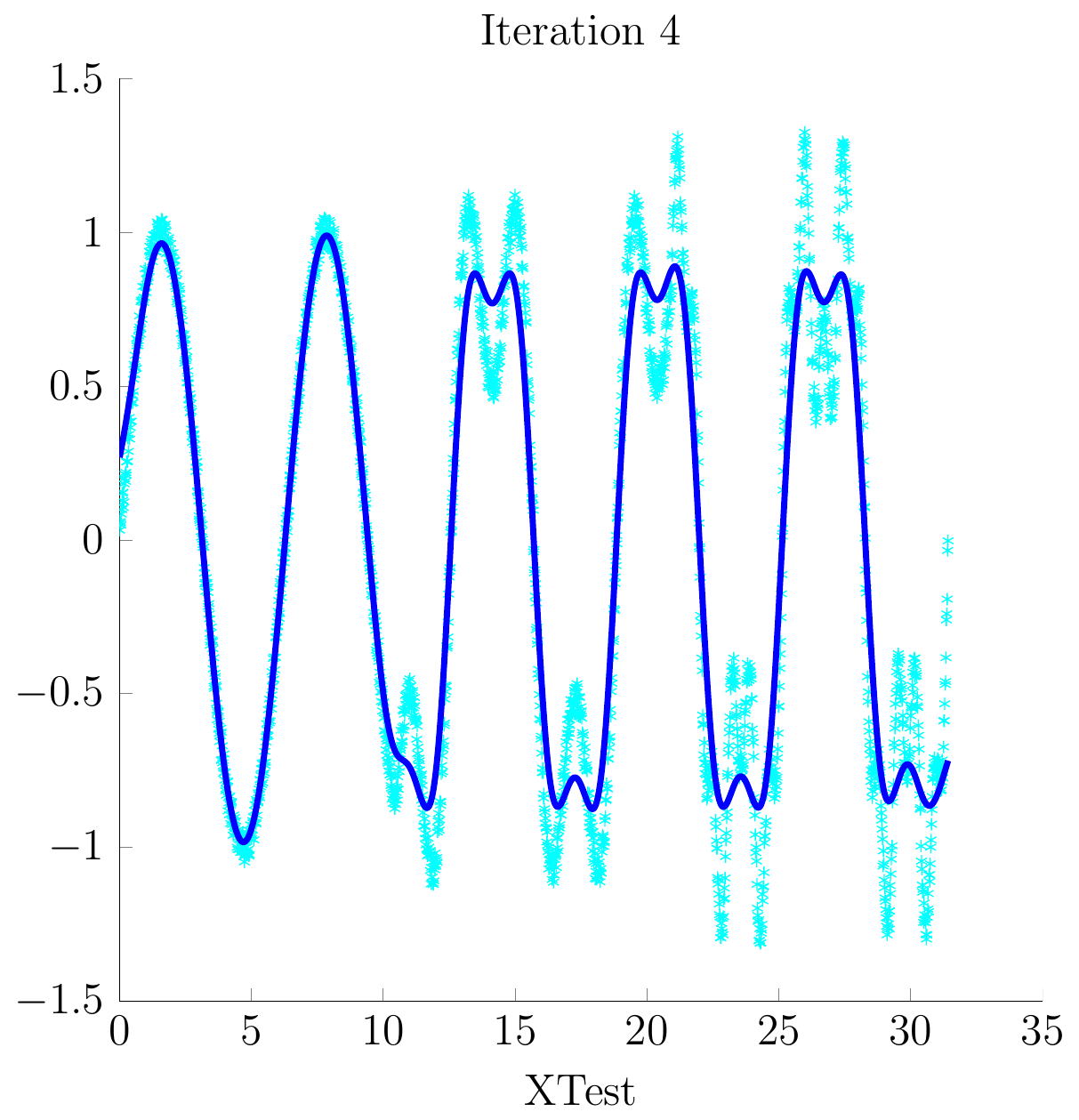}} \\
	\subfloat{\includegraphics[width=0.24\textwidth]{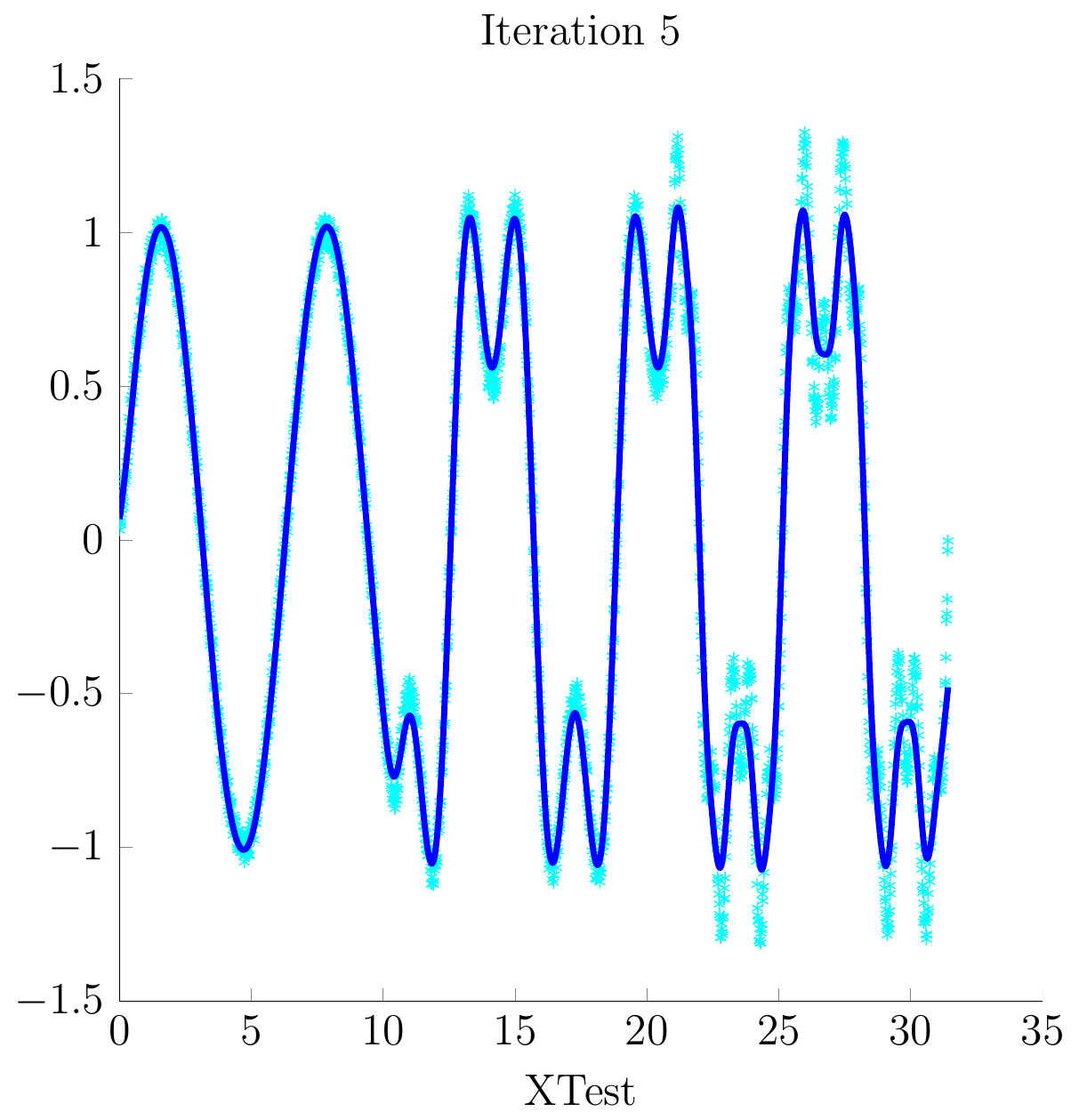}}
	\subfloat{\includegraphics[width=0.24\textwidth]{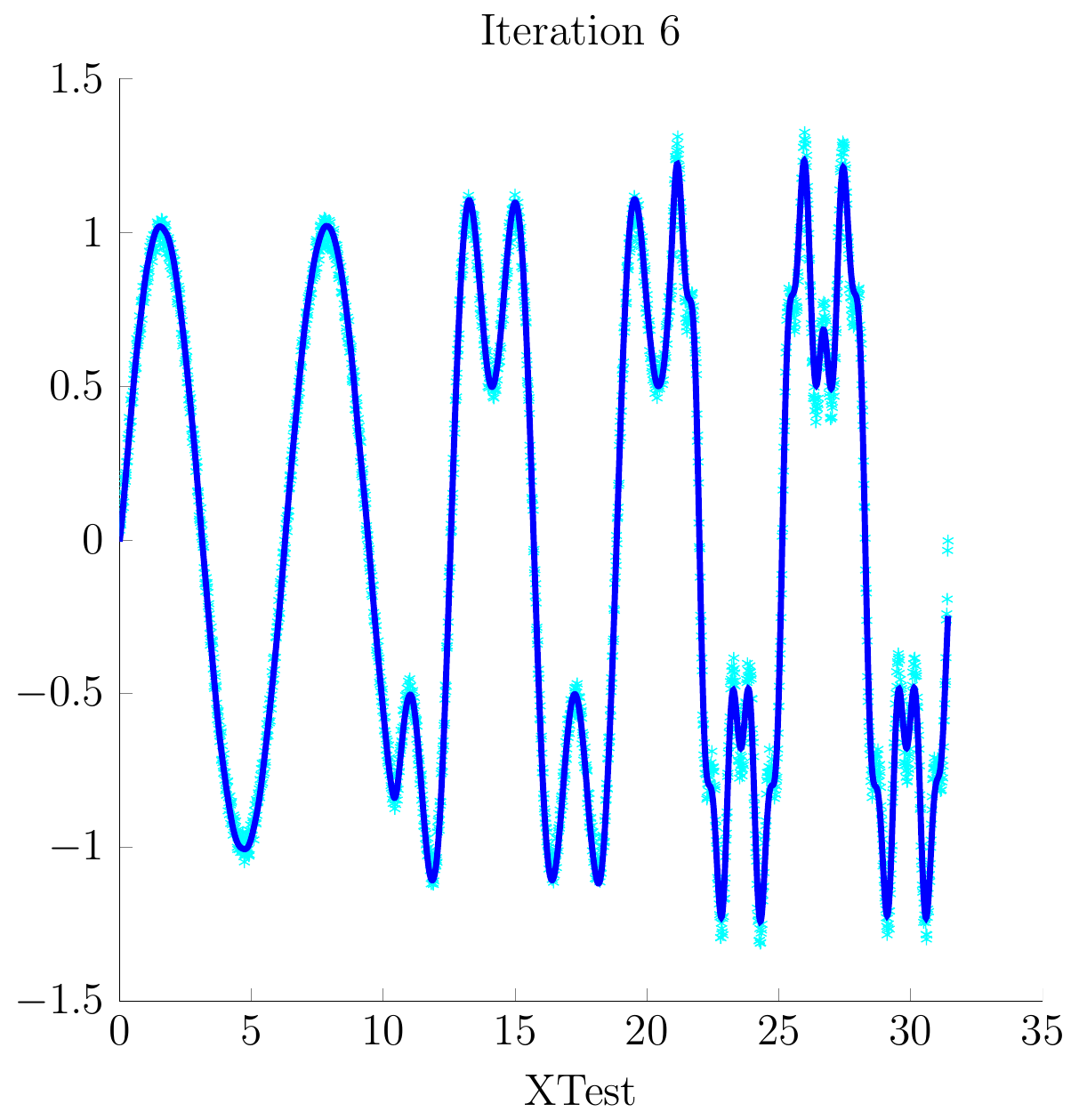}}
	\subfloat{\includegraphics[width=0.24\textwidth]{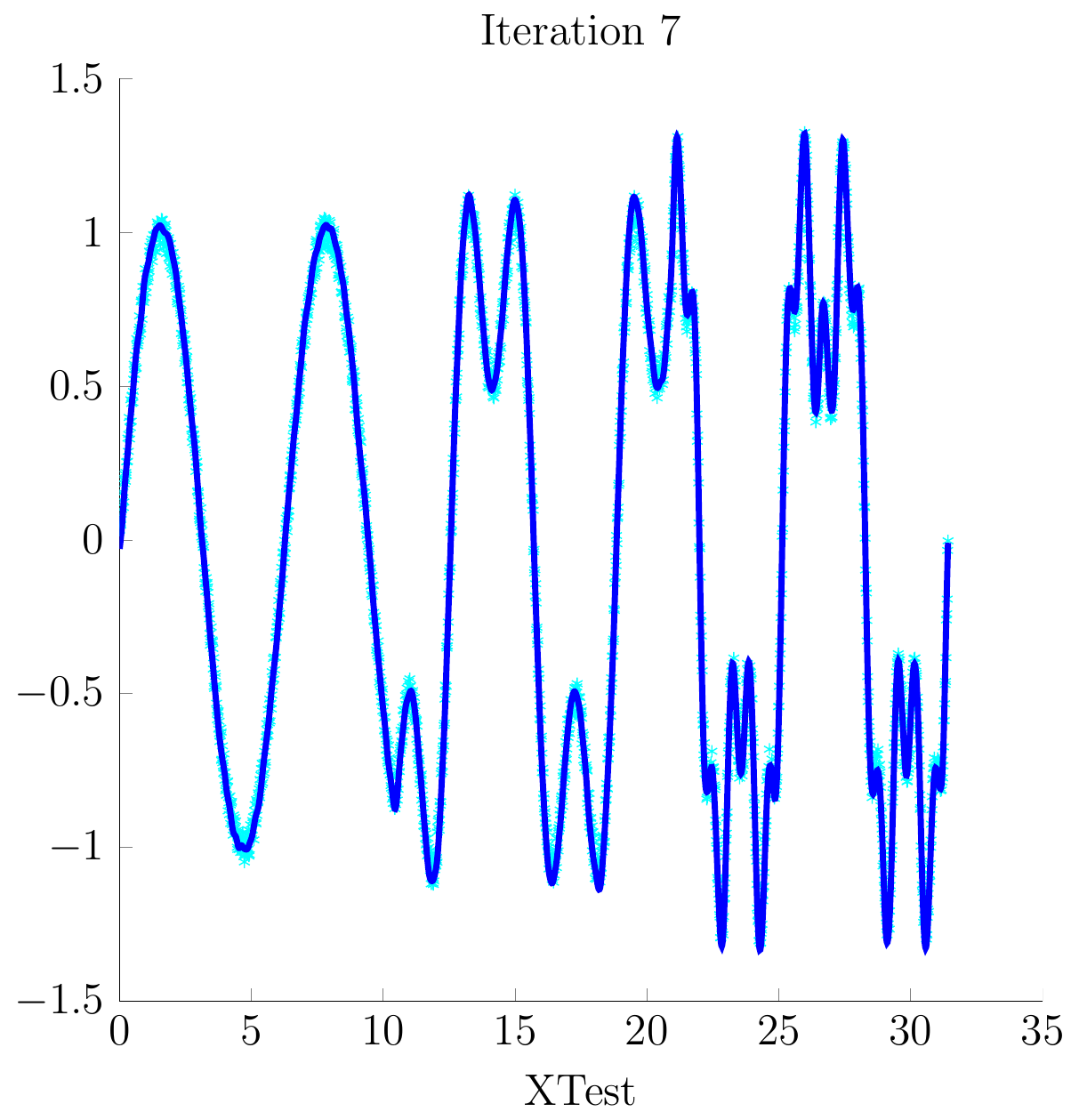}}
\caption{Evolution of the Auto-adaptative Laplacian Pyramids model for the first example.} 
\label{behavior} 
\end{figure*}

\section{A Synthetic Example}
\label{sec:Synt}

For a better understanding of this theory, we first illustrate the proposed ALP algorithm on a synthetic example of a composition  of sines with different frequencies plus additive noise. 

We consider a sample $x$ with $N$ points equally spaced over the range $[0, 10\pi]$.
The target function $f$ is then
\begin{equation*}
f = \sin(x) +0.5\sin(3x) \cdot I_2(x) +0.25\sin(9x) \cdot I_3(x) +\varepsilon,
\end{equation*}
where $I_2$ is the indicator function of the interval $(10\pi/3, 10\pi]$, $I_3$ that of $(2 \cdot 10\pi/3, 10\pi]$ and $\varepsilon \sim \mathcal{U}([-\delta,\delta])$ is uniformly distributed noise.
In other words, we have a single frequency in the interval $[0, 10\pi/3]$, two frequencies in $(10\pi/3, 2 \cdot 10\pi/3]$ and three in $(2 \cdot 10\pi/3, 10\pi]$.
We run two different simulations, the first one with $4,000$ points with small $\delta = 0.05$ noise and the second one with $2,000$ points and a larger $\delta = 0.25$ (observe that $|f| \leq 1.75$).
In both cases, odd indexed points form the training set and even points form the test set.

Recall that the ALP model automatically adapts its multiscale behavior to the data, trying to refine the prediction in each iteration using a more localized kernel, given by a smaller $\sigma$. This behavior can be observed in Figure~\ref{behavior}, which shows the evolution of the prediction of the ALP for small noise experiment.
As we can see, at the beginning, the model approximates the function just by a coarse mean of the target function values; however in the subsequent iterations that start using sharper kernels and refined residuals, the approximating function starts capturing the different frequencies and amplitudes of the composite sines.
In this particular case the minimum LOOCV value is reached after $7$ iterations, a relatively small number which makes sense as we have a simple model with small noise. 
\begin{figure}[ht!]
\centering
	\includegraphics[width=0.45\textwidth]{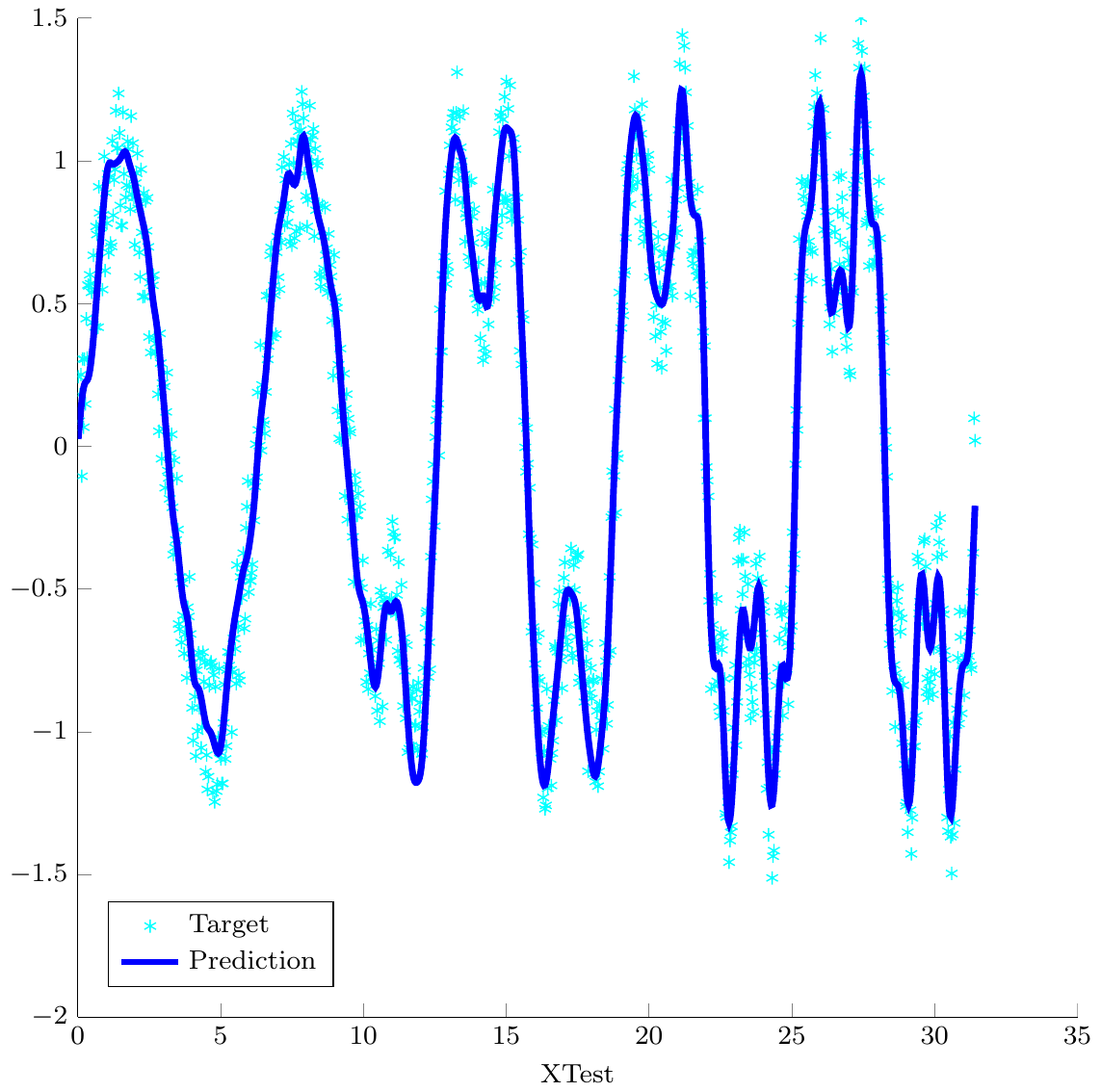}
\caption{Prediction of the Auto-adaptative Laplacian Pyramids model for the second example.} 
\label{syn2} 
\end{figure}

When we repeat the same synthetic experiment but now enlarging the amplitude of the uniform distribution to $\delta = 0.25$, the predicted function is represented in Figure~\ref{syn2} and it is obtained after $6$ iterations. 
As it was to be expected, the number of LP iterations is now slightly smaller than in the previous example because the algorithm selects a more conservative, smoother prediction in the presence of noisier and, thus, more difficult data.

In any case, we can conclude that the ALP model captures very well the essential behavior underlying both samples, catching the three different frequencies of the sine and their amplitudes even when the noise level increases. 

\section{ALP for Eigenvector Estimation in Spectral Clustering and Diffusion Maps}
\label{sec:Real}

\subsection{Spectral Manifold Learning}

A common assumption in many big data problems is that although original data appear to have a very large dimension, they lie in a low dimensional manifold ${\cal M}$ of which a suitable representation has to be given for adequate modeling.
How to identify ${\cal M}$ is the key problem in manifold learning, where the preceding assumption has given rise to a number of methods among which we can mention Multidimensional Scaling~\cite{MDSCox}, Local Linear Embedding~\cite{LLERoweis}, Isomap~\cite{IsomapTenenbaum}, Spectral Clustering (SC)~\cite{SCShi} and its several variants such as Laplacian Eigenmaps~\cite{LEBelkin}, or Hessian eigenmaps~\cite{HEDonoho}.
Diffusion methods (DM)~\cite{DMCoifman} also follow this set up and we center our discussion on them. 

The key assumption in Diffusion methods is that the metric of the low dimensional Riemannian manifold where data lie can be approximated by a suitably defined diffusion metric.
The starting point in DM (and in SC) is a weighted graph representation of the sample with a similarity matrix $w_{ij}$ given by the kernel $e^{-\sfrac{\|x_i-x_j\|^2}{2\sigma^2}}$.
In order to control the effect of the sample distribution, a parameter $\alpha \in [0,1]$ is introduced in DM and we work instead with the $\alpha$-dependent similarity $w^{(\alpha)}_{ij} = \sfrac{w_{ij}}{g_i^{\alpha} g_j^{\alpha}}$, where $g_i = \sum_j w_{ij}$ is the degree of vertex $i$.
The new vertex degrees are now $g^\alpha_i = \sum_j w^\alpha_{ij}$ and we define a Markov transition matrix $\tilde{W}^{(\alpha)}=\{\tilde{w}^{(\alpha)}_{ij} = \sfrac{w^{(\alpha)}_{ij}}{g^\alpha_i}\}$.
Fixing the number $t$ of steps considered in the Markov process, we can take into account $t$ neighbors of any point in terms of the $t$-step diffusion distance given by 
\begin{equation*}
D^{t}_{ij} =  \| \tilde{w}^{(\alpha;t)}_{i,\cdot} - \tilde{w}^{(\alpha;t)}_{j,\cdot} \|_{L^2(\sfrac{1}{\phi_{0}})},
\end{equation*}
where $\phi_{0}$ is the stationary distribution of the Markov process and $\tilde{w}^{(\alpha;t)}$ the transition probability in $t$ steps (i.e., the $t$-th power of $\tilde{w}^{\alpha}$).
We will work with $t=1$ and use $\alpha=1$.
As it is shown in~\cite{DMCoifman},  when $\alpha=1$, the infinitesimal generator $L_1$ of the diffusion process coincides with the manifold's Laplace--Beltrami operator $\Delta$ and, thus, we can expect the diffusion projection to capture the underlying geometry.

Now, the eigenanalysis of the Markov matrix $\tilde{W}^{(\alpha)}$ gives an alternative representation of the diffusion distance as 
\begin{equation*}
D^{t}_{ij} = \sum_k \lambda_k^{2t} (\psi^{(k)}_{i} - \psi^{(k)}_{j})^2,
\end{equation*}
with $\lambda_k$ the eigenvalues and $\psi^{(k)}$ the left eigenvectors of the Markov matrix $\tilde{W}^{(\alpha)}$ and $\psi^{(k)}_{i} = \psi^k (x_i)$ are the eigenvectors' components~\cite{DMCoifman}.
The eigenvalues $\lambda_k$ decay rather fast, a fact we can use to perform dimensionality reduction by setting a fraction $\delta$ of the second eigenvalue $\lambda_1$ (the first one $\lambda_0$ is always $1$ and carries no information) and retaining thus a number $d = d_t$ of those $\lambda_k$ for which $\lambda^t_k > \delta \lambda^t_1$.
This $\delta$ is a parameter we have to choose besides the previous $\alpha$ and $t$. Usual values are either $\delta=0.1$ or the more strict (and larger dimension inducing) $\delta=0.01$.
Once fixed, we would thus arrive to the diffusion coordinates 
\begin{equation*}
\Psi = \begin{pmatrix} \lambda_{1}^t \psi_{1}(x) \\ \vdots \\ \lambda_{d}^t \psi_{d}(x) \end{pmatrix}
\end{equation*}
and we can approximate the diffusion distance as
\begin{equation*}
D^{t}_{ij} \sim \sum_1^d \lambda_k^t (\psi^k_{i} - \psi^k_{j})^2 = \| \Psi(x_i) - \Psi(x_j) \|^2.
\end{equation*}
In other words, the diffusion distance in ${\cal M}$ can be approximated by Euclidean distance in the DM projected space. 
This makes DM a very useful tool to apply procedures in the projected space such as $K$-means clustering that usually rely on an implicit Euclidean distance assumption.
All the steps to compute DM are summarized in Algorithm \ref{alg:DM}.
\begin{algorithm} [ht!]
\caption{Diffusion Maps Algorithm.}
 \begin{algorithmic}[1]
  \REQUIRE ${\cal S}=\{x_1, \dots, x_N\}$, the original dataset. \\ Parameters: $t$, $\alpha$, $\sigma$, $\delta$.
  \ENSURE $\{\Psi(x_1), \dots, \Psi(x_N)\}$, the embedded dataset.
  
  \STATE Construct a graph $G=({\cal S}, W)$ where $$W_{ij}=w(x_i,x_j)=e^{\frac{-\|x_i - x_j\|^2}{2\sigma^2}}.$$
  \STATE Define the initial density function as $$g(x_{i}) = \sum_{j=1}^{N} w(x_{i}, x_{j}).$$
  \STATE Normalize the weights by the density: $$w^{(\alpha)}(x_i, x_j) = \frac{w(x_i, x_j)}{g_i^{\alpha} g_j^{\alpha}}.$$
  \STATE Define the transition probability $$\tilde{W}^{(\alpha)}_{ij}=\tilde{w}^{(\alpha)}(x_i,x_j) = \frac{\tilde{w}^{(\alpha)}(x_i,x_j)}{g^{(\alpha)}_i},$$
\\ where $g^{(\alpha)}_i = \sum_{j=1}^{n} w^{(\alpha)}(x_i,x_j)$ is the graph degree.
  \STATE Obtain eigenvalues $\{\lambda_{r}\}_{r \geqslant 0}$ and eigenfunctions $\{\psi_{r}\}_{r \geqslant 0}$ of $\tilde{W}^{(\alpha)}$ such that $$\left \{ \begin{array}{rcl} 1 & = & \lambda_0 > |\lambda_1| \geqslant \cdots \\ \tilde{W}^{(\alpha)} \psi_{r} & = & \lambda_{r} \psi_{r} . \end{array} \right.$$
  \STATE Compute the embedding dimension using a threshold $d = \max \{ \ell : |\lambda_{\ell}| > \delta |\lambda_{1}| \}$.
  \STATE Formulate Diffusion Map: $$\Psi = \begin{pmatrix} \lambda_{1}^t \psi_{1}(x) \\ \vdots \\ \lambda_{d}^t \psi_{d}(x) \end{pmatrix}.$$
 \end{algorithmic}
\label{alg:DM}
\end{algorithm}

While very elegant and powerful, DMs have the drawback of relying on the eigenvectors of the Markov matrix. This makes it difficult to compute the DM coordinates of a new, unseen pattern $x$. Moreover, the eigenanalysis of the $\tilde{W}^{(\alpha)}$ matrix would have in principle a potentially very high $O(N^3)$ cost, with $N$ sample's size.
However, both issues can be addressed in terms of function approximation. The standard approach is to apply the Nystr\"{o}m extension formula \cite{NysWilliams} but LPs \cite{LPRabin} and, hence, ALPs, can also be used in this setting, as we discuss next.

To do so we consider the eigenvalue components $\psi^{(k)}_{j}$ as the values $\psi^{(k)} (x_j)$ at the points $x_j$ of an unknown function $\psi^{(k)}(x)$ which we try to approximate by an ALP scheme.
The general LP formula \eqref{LPmodel} for the eigenvector $\hat{\psi}^{(k)}$ extended to an out-of-sample point $x$ becomes now 
\begin{align}
	\hat {\psi}^{(k)} (x) & = \psi^{(k)} * P^0 (x) + \sum_0^{L-1} d_{\ell} * P^{\ell+1} (x) \nonumber \\
& = \sum_{i=1}^{N} {\cal P}_0(x, x_i) {\psi}^{(k)}(x_i) + \sum_{h=1}^{H} \sum_{i=1}^{N} {\cal P}_h(x,x_i) d^{(k)}_{h}(x_i),
\label{DMALP}
\end{align}
with the differences $d^{(k)}_{h}$ given now by $d^{(k)}_{h} (x_i) = \psi^{(k)}(x_i) - \sum_{\ell=0}^{h-1} \hat ({\psi}^{(k)})^{(\ell)}(x_i)$.

We illustrate next the application of these techniques to the analysis of solar radiation data where we relate actual aggregated radiation values with Numerical Weather Prediction (NWP) values.

\subsection{Diffusion Maps of Radiation Data}
A current important problem that is getting a growing attention in the Machine Learning community is the prediction of renewable energies, particularly solar energy and, therefore, of solar radiation. 
We will consider here the prediction of the total daily incoming solar energy in a series of meteorological stations located in Oklahoma in the context of the AMS 2013-2014 Solar Energy Prediction Contest hosted by the Kaggle company.\footnote{American Meteorological Society 2013-2014 Solar Energy Prediction Contest (\url{https://www.kaggle.com/c/ams-2014-solar-energy-prediction-contest}).}
While the ultimate goal would be here to obtain best predictions, we will use the problem to illustrate the application of ALP in the previously described DM setting. 

The input data are an ensemble of 11 numerical weather predictions (NWP) from the NOAA/ESRL Global Ensemble Forecast System (GEFS).
We will just use the main NWP ensemble as being the one with the highest probability of being correct.
Input patterns contain five time-steps (from $12$ to $24$ UCT-hours in $3$ hour increments) with $15$ variables per time-step for all points in a $16\times9$; each pattern has thus a very large $10,800$ dimension. 
The NWP forecasts from 1994--2004 yield $4,018$ training patters and the years 2005, 2006 and 2007, with $1,095$ patterns, are used for testing.

Our first goal is to illustrate how applying ALP results in good approximations to the DM coordinates of new points that were not available for the first eigenanalysis of the initial Markov matrix.
To do so, we normalize training data to $0$ mean and a standard deviation $1$ and use the DM approach explained above, working with a Gaussian kernel, whose $\sigma$ parameter has been established as the $50\%$ percentile of the Euclidean distances between all sample points. 
We recall that we fix the diffusion step $t$ to $1$ and also the $\alpha$ parameter, so that data density does not influence diffusion distance computations.
To decide the best dimensionality for the embedding, the precision parameter $\delta$ has been fixed at a relatively high value of $0.1$, i.e., we only keep the eigenvalues that are bigger than the $10\%$ of the first non trivial eigenvalue. 
This choice yields an embedding dimension of $3$, which enables to visualize the results.
We apply DM with these parameters over the training set and obtain the corresponding eigenvectors, i.e., the sample values $\psi^{(k)}(x_i)$ over the training patterns $x_i$, and the DM coordinates of the training points.
We then apply Algorithm~\ref{alg:train} to decide on the ALP stopping iteration and to compute the differences $d^{(k)}_{h}(x_i)$ in \eqref{DMALP} and finally apply Algorithm~\ref{alg:test} to obtain the approximations to the values of $\psi^{(k)}$ over the test points.

In order to measure the goodness of the new coordinates, we have also performed the DM eigenanalysis to the entire NWP data, i.e., to the training and test inputs together.
In this way we can compare the extended embedding obtained using ALP with the one that would have been obtained if we had known the correct diffusion coordinates.

\subsection{Results}

\begin{figure*}[ht!]
\centering
	\subfloat[Training Radiation.]{\includegraphics[width=0.245\textwidth]{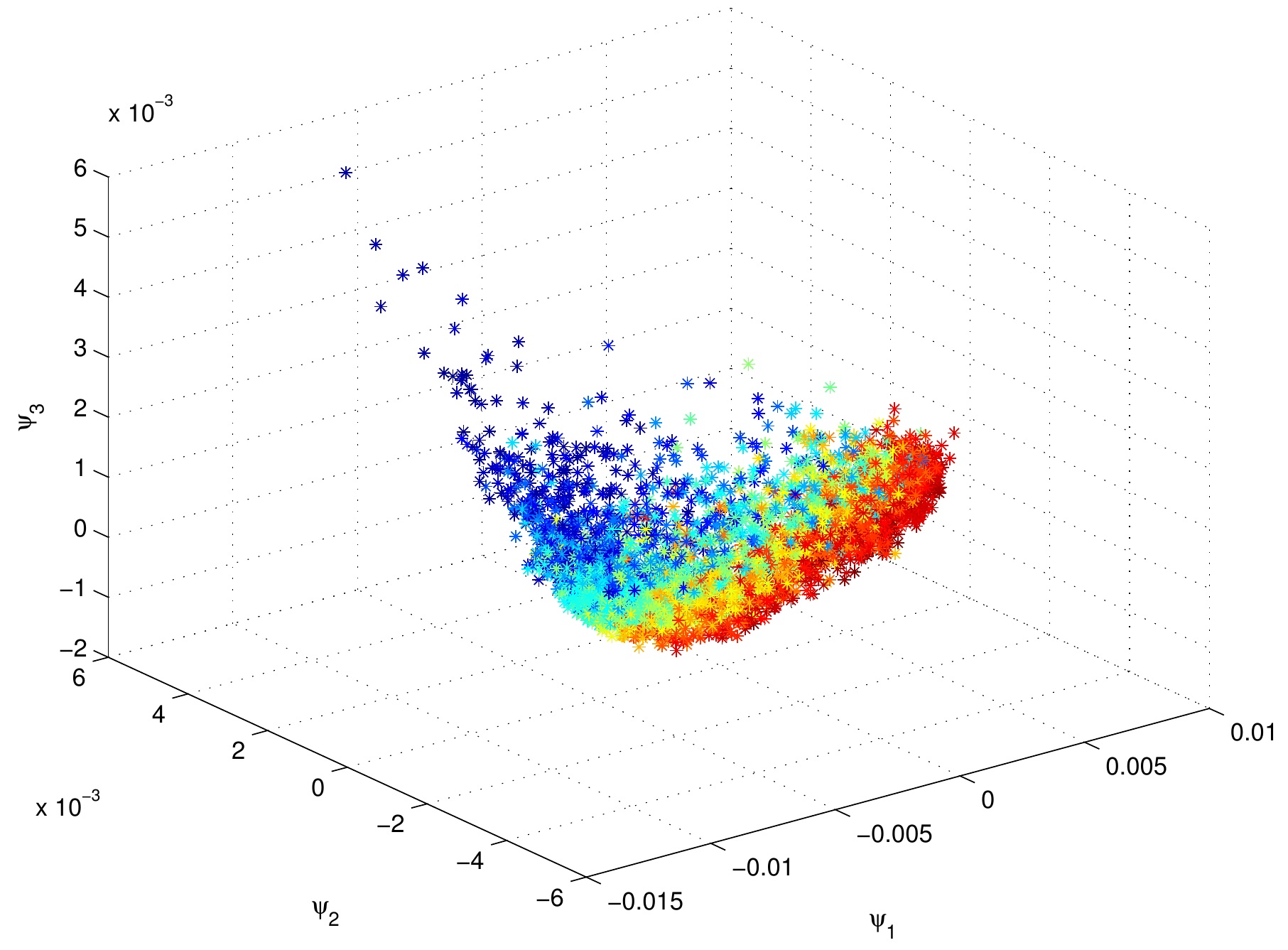}} 
	\subfloat[Training ALP Prediction.]{\includegraphics[width=0.245\textwidth]{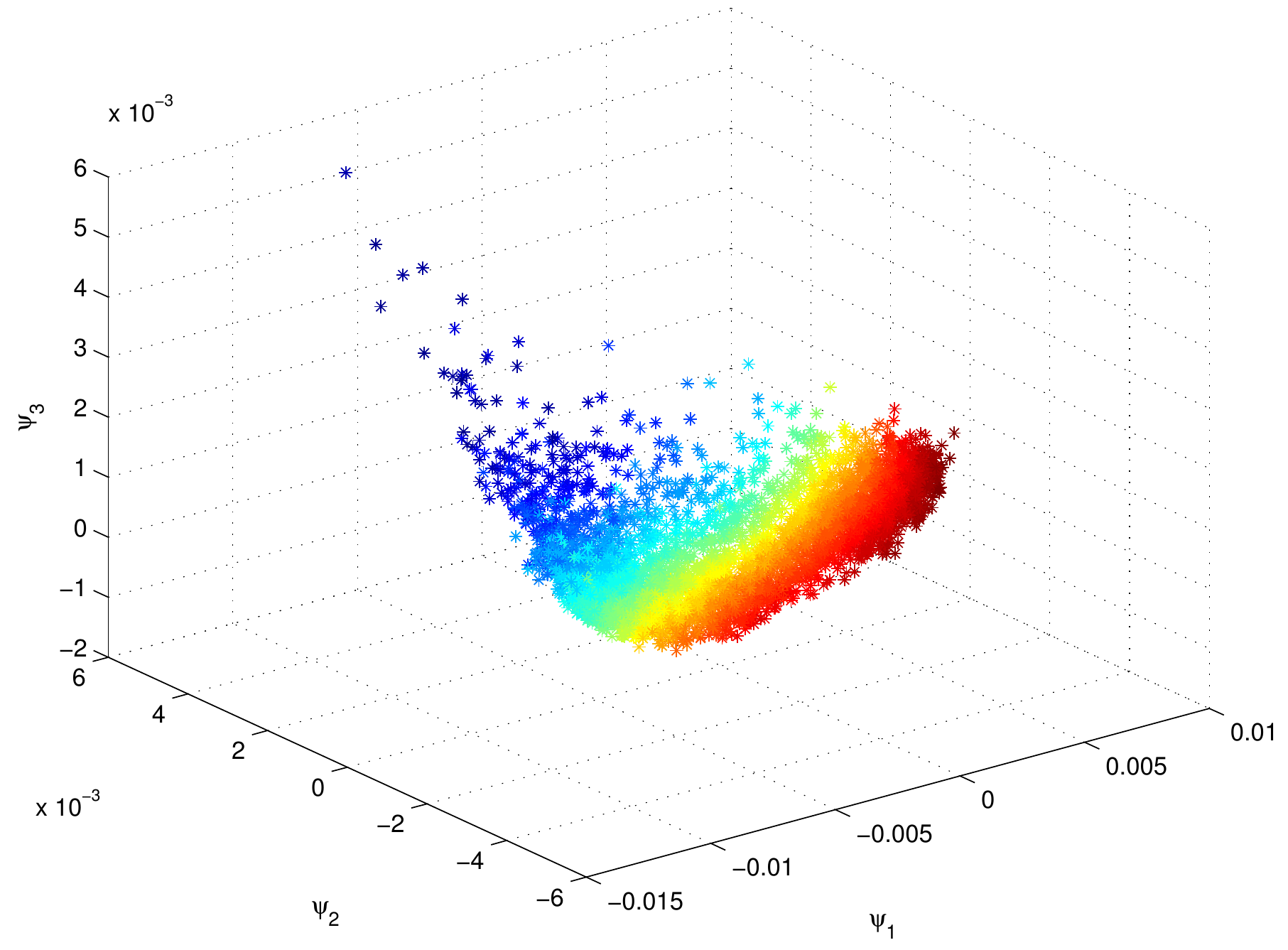}}
	\subfloat[Test Radiation.]{\includegraphics[width=0.245\textwidth]{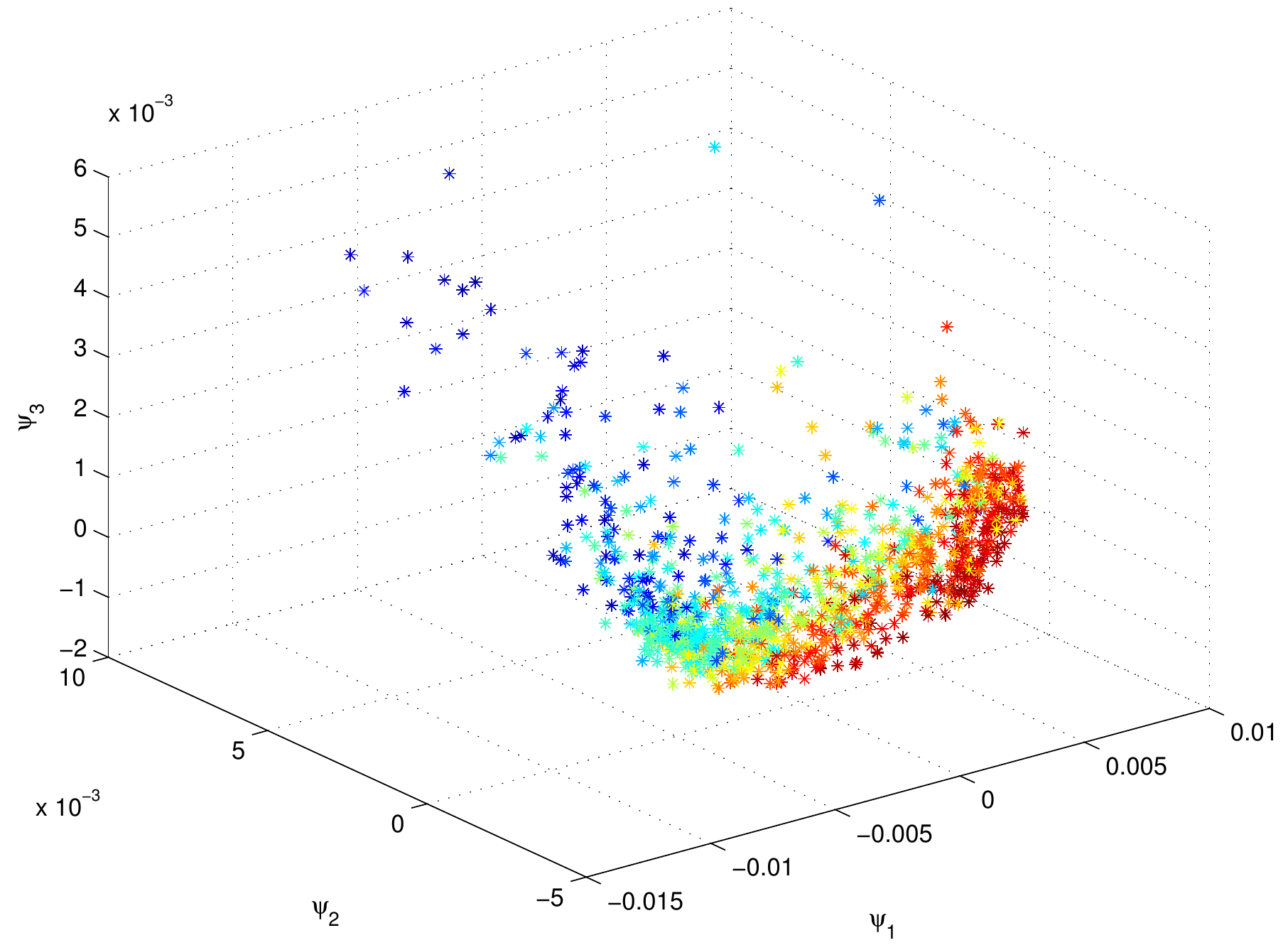}}
	\subfloat[Test ALP Prediction.]{\includegraphics[width=0.245\textwidth]{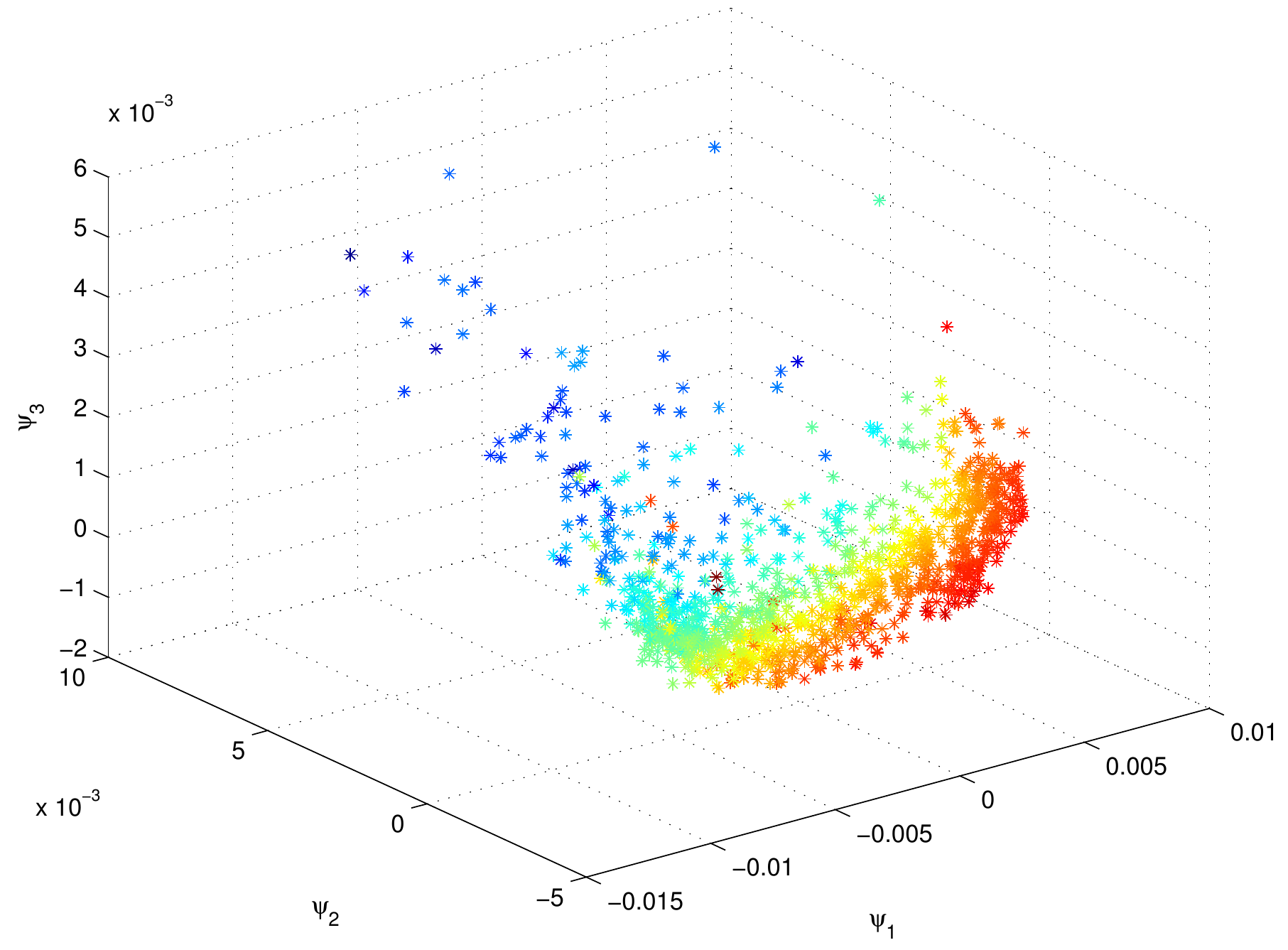}}
\caption{Training and test results when the DM embedding is computed only on the training sample and ALP is used first to extend DM to the test sample and then for radiation prediction.}
\label{embeddingLP}
\end{figure*}

For this experiments the results have been obtained computing a DM embedding only over the training sample, and using ALP first to extend DM to the test sample and then for radiation prediction. In Figure~\ref{embeddingLP} training and test results are shown. 
The three diffusion coordinates for this example are colored by the target, i.e. by the solar radiation (first and third), or by the prediction given by ALP (second and fourth).
\begin{figure*}[ht!]
\centering
	\subfloat[Training Radiation.]{\includegraphics[width=0.245\textwidth]{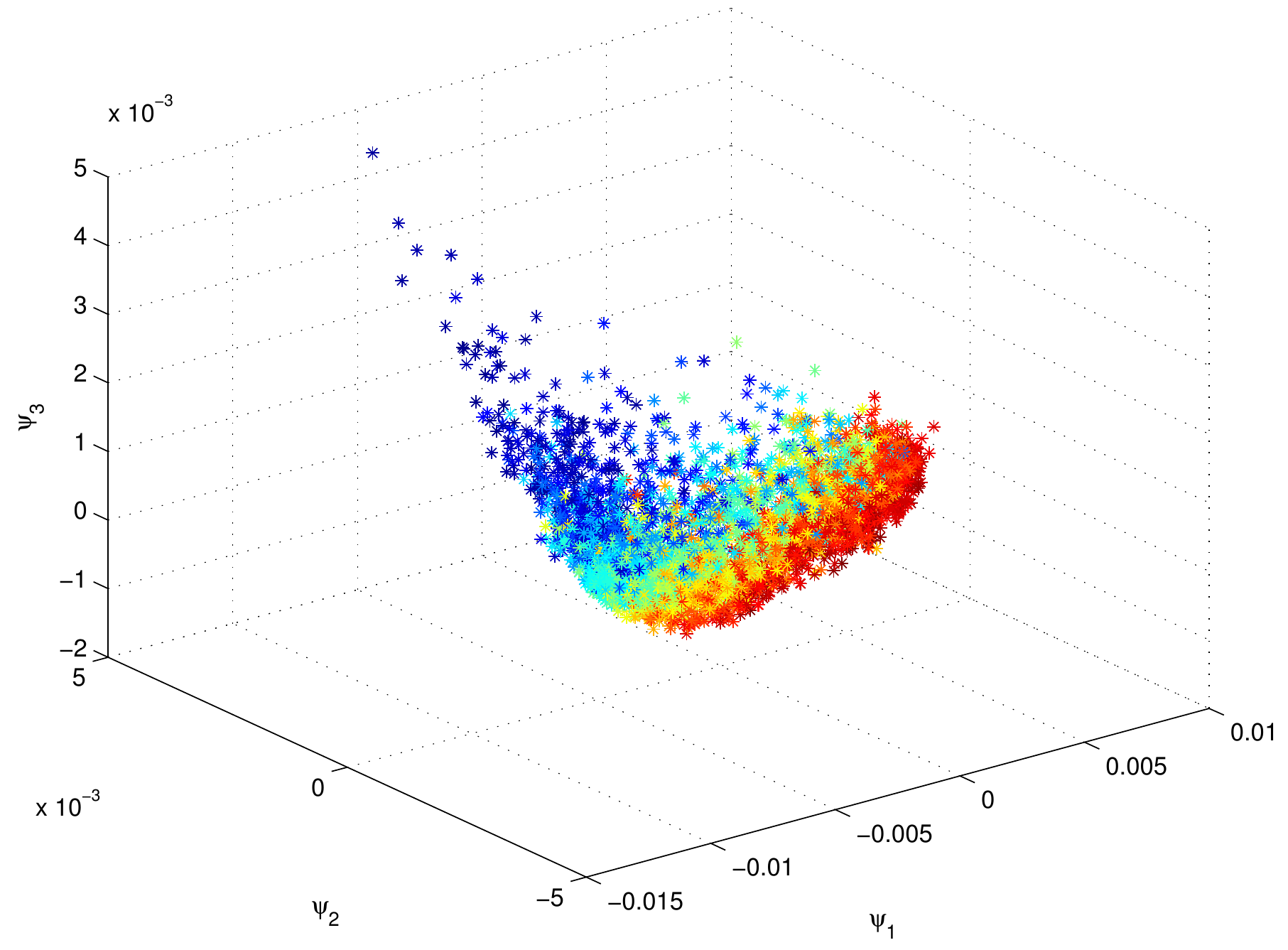}} 
	\subfloat[Training ALP Prediction.]{\includegraphics[width=0.245\textwidth]{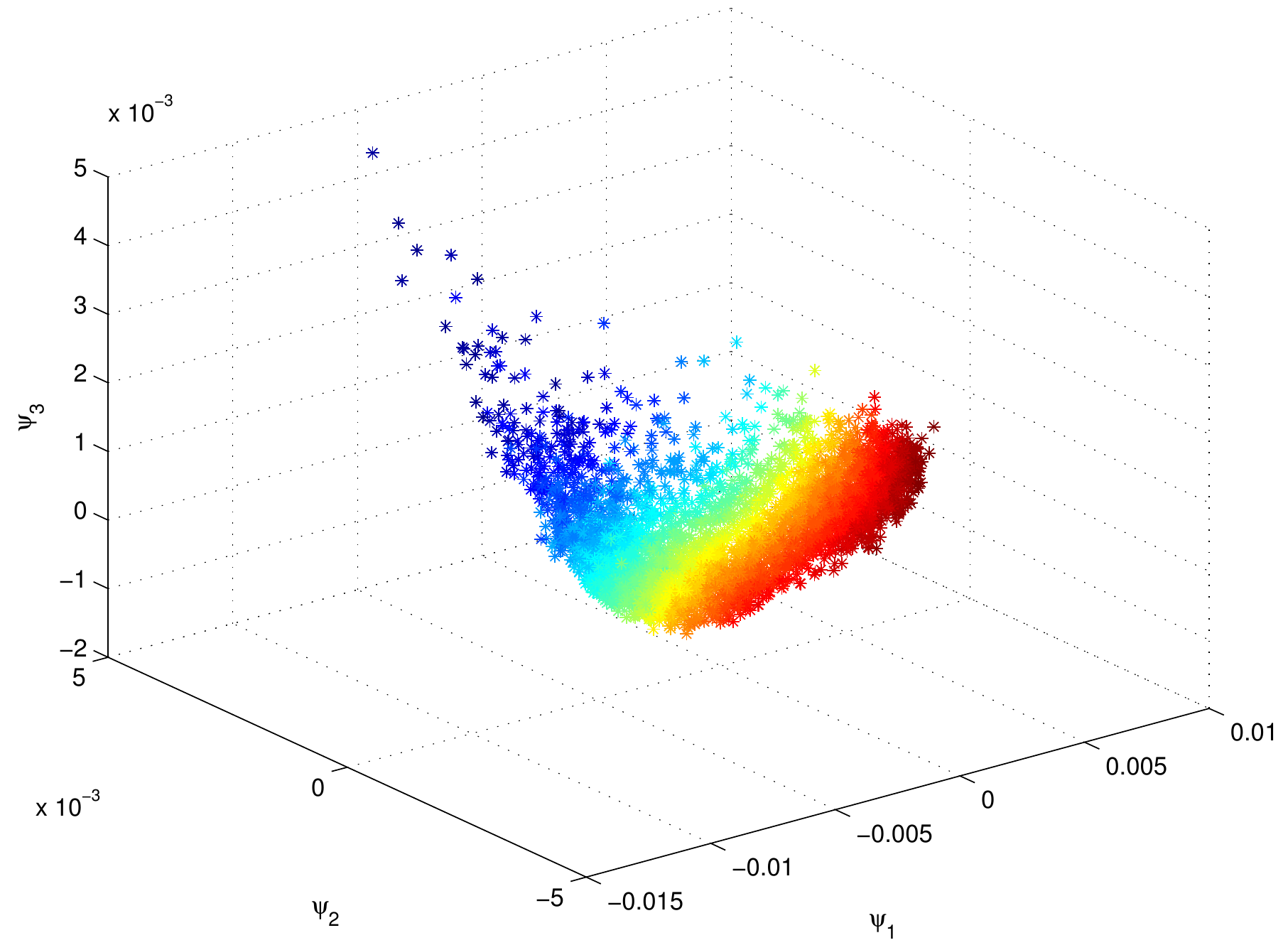}}
	\subfloat[Test Radiation.]{\includegraphics[width=0.245\textwidth]{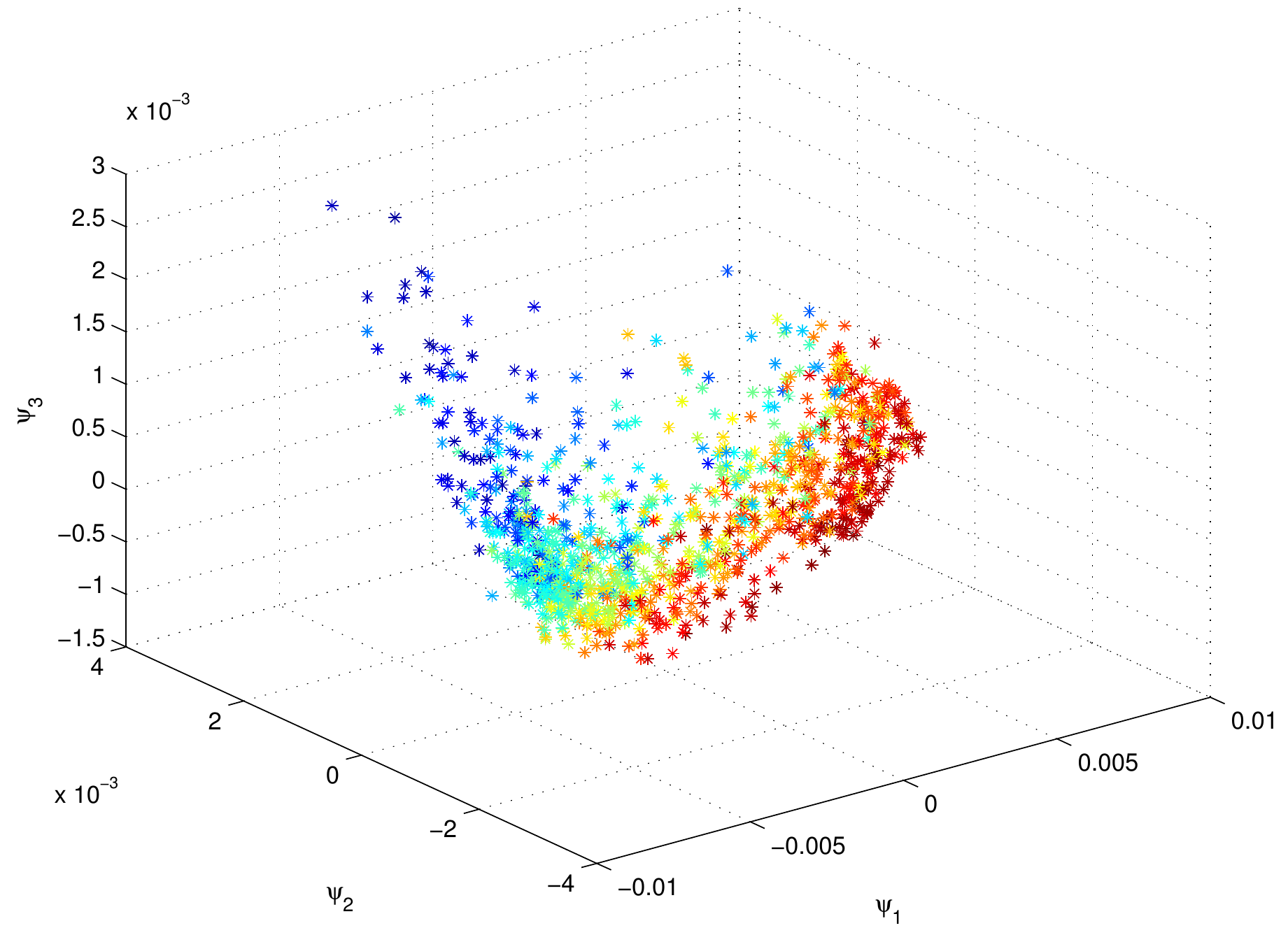}}
	\subfloat[Test ALP Prediction.]{\includegraphics[width=0.245\textwidth]{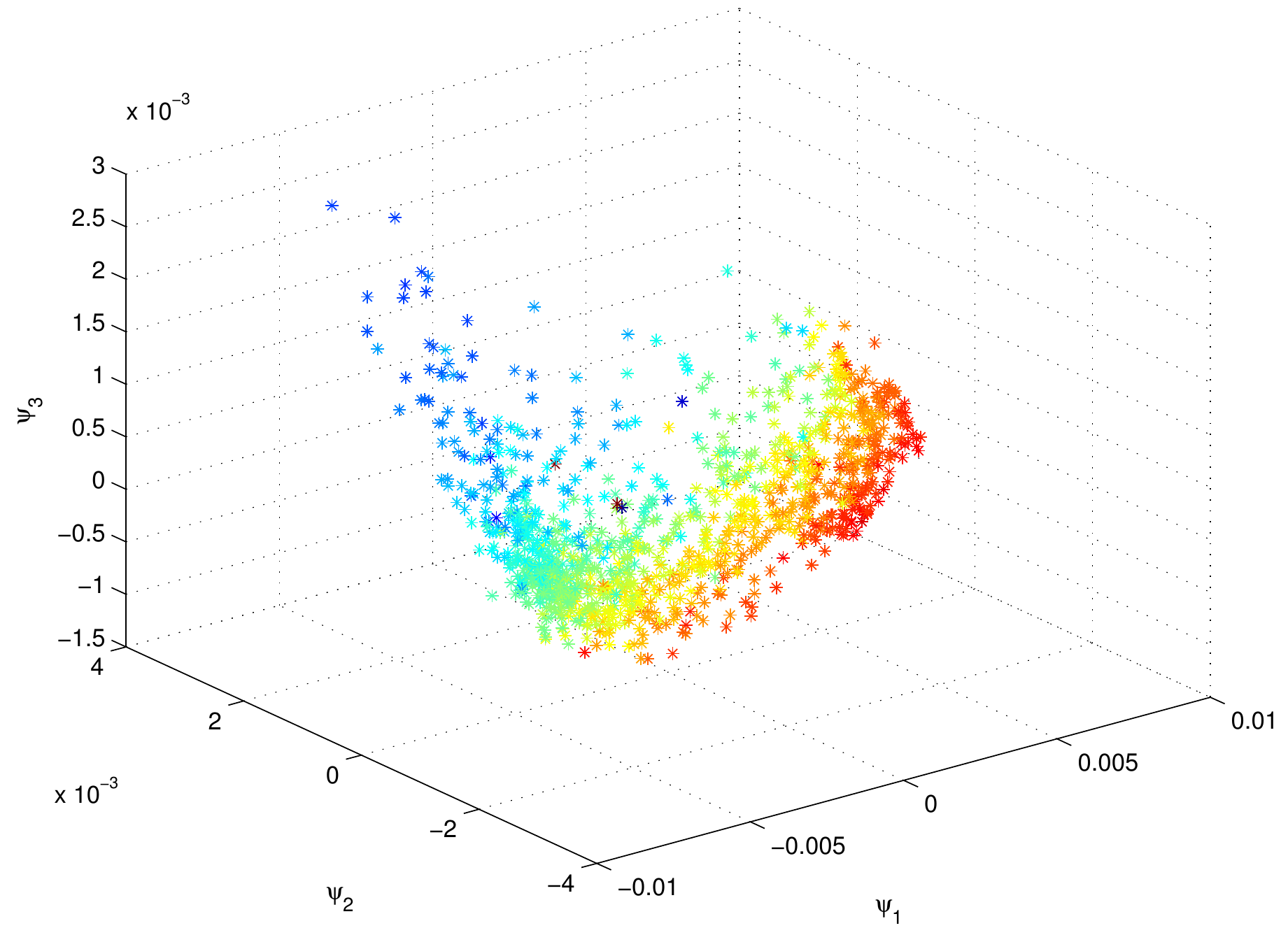}}
\caption{Training and test results when the DM embedding is computed over the entire sample and ALP used only for radiation prediction.}
\label{embedding}
\end{figure*}
\begin{figure*}[ht!]
\centering
	\subfloat[Training DM Full.]{\includegraphics[width=0.245\textwidth]{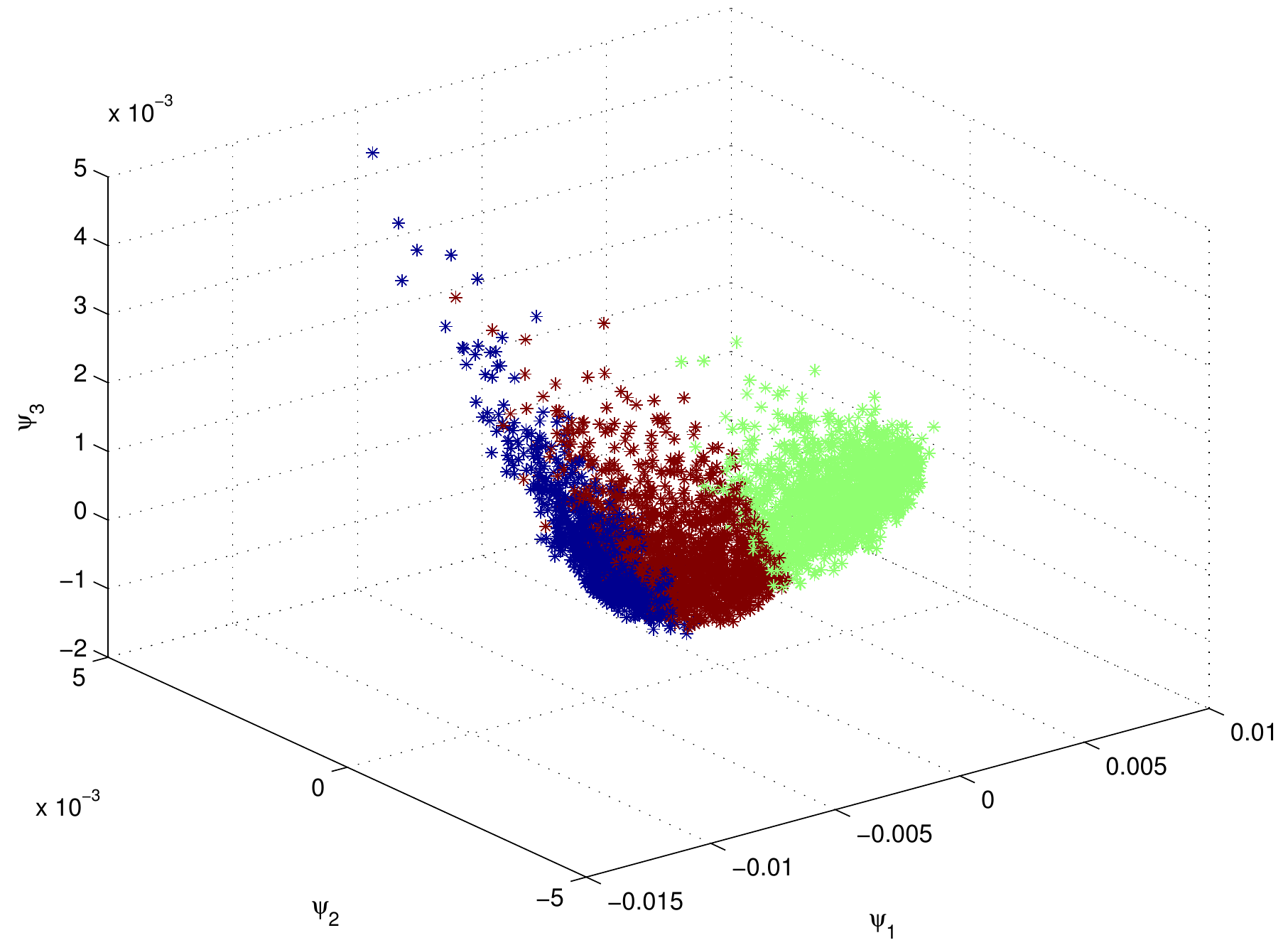}} 
	\subfloat[DM Training Sample.]{\includegraphics[width=0.245\textwidth]{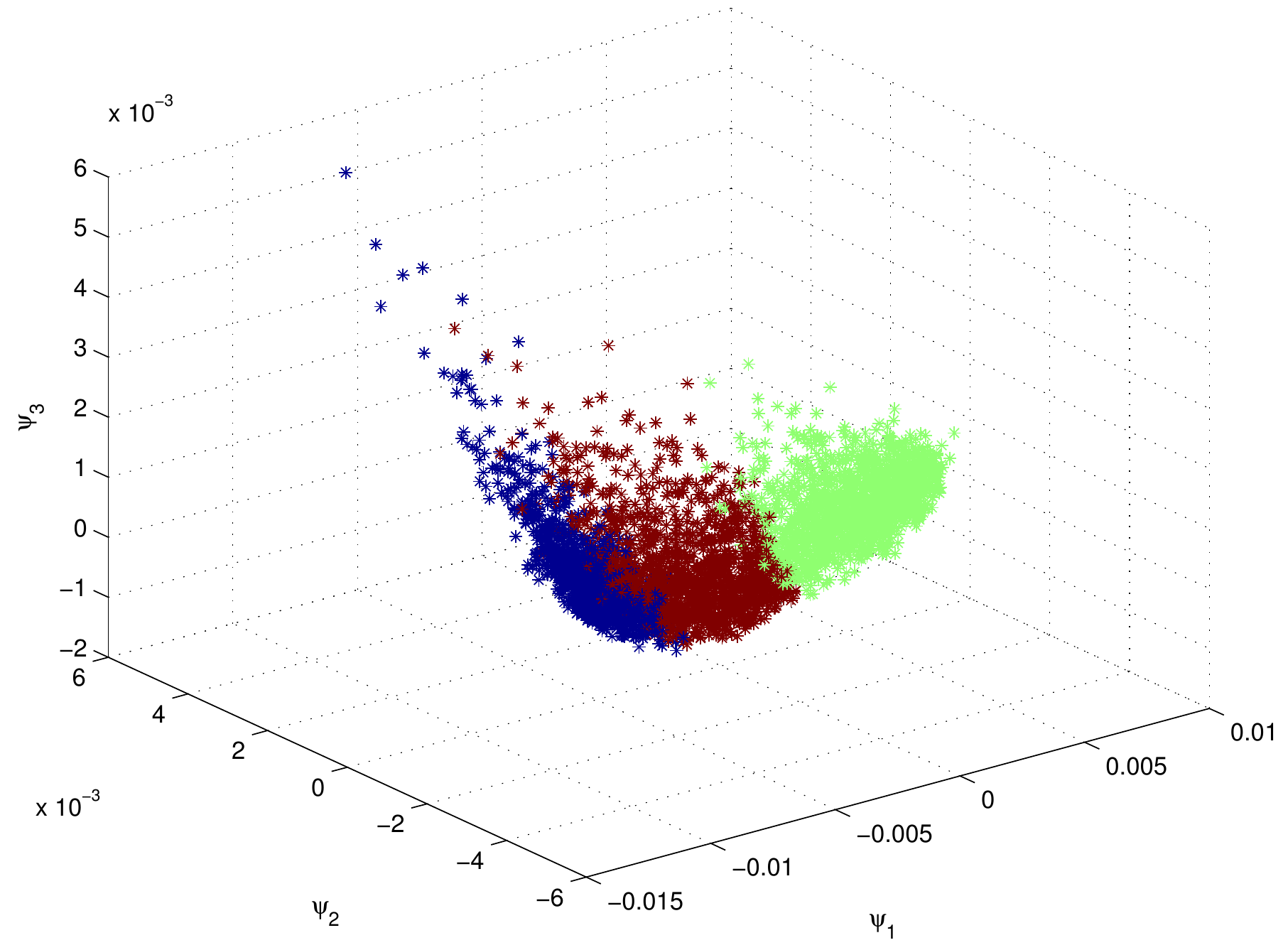}}
	\subfloat[Test DM Full.]{\includegraphics[width=0.245\textwidth]{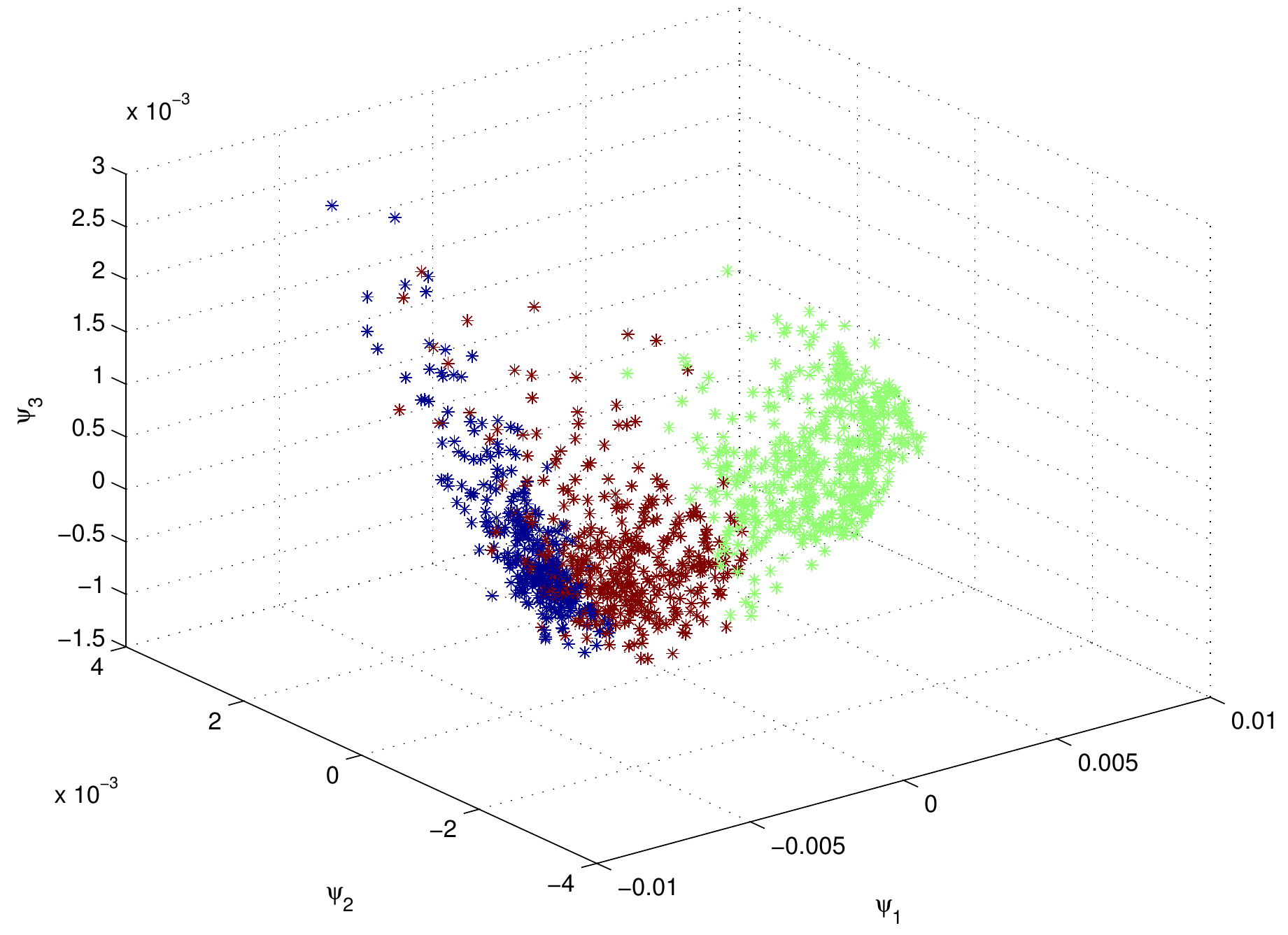}}
	\subfloat[Test Expanded by ALP.]{\includegraphics[width=0.245\textwidth]{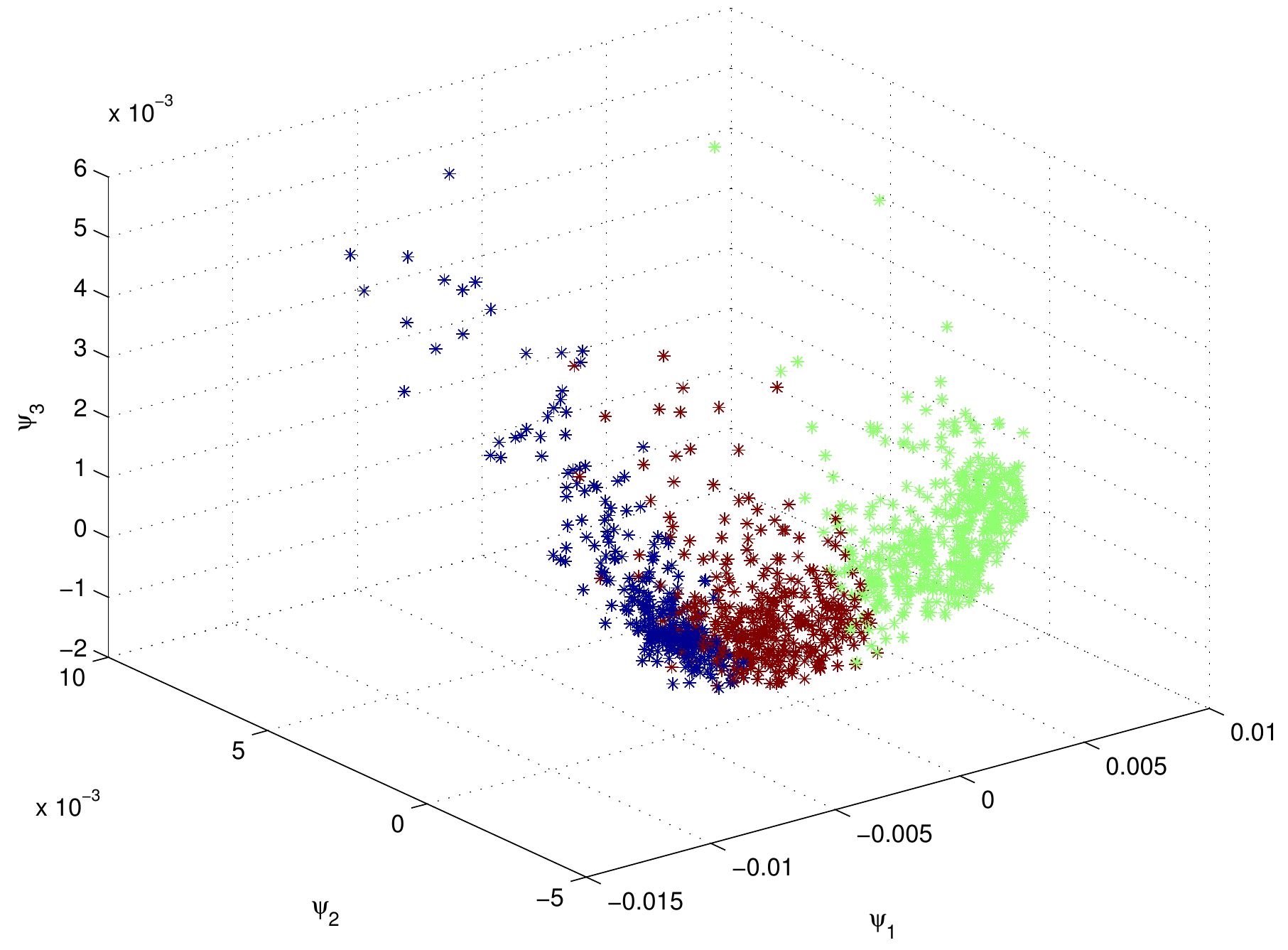}}
\caption{Clusters over the embedded data.}
\label{clusters}
\end{figure*}
In the first and third image we can see that DM captures the structure of the target radiation in the sample, with the low radiation data (blue) appearing far apart from the points with high radiation (red). 
If we compare this with the prediction-colored embeddings (second and fourth) we can observe that the radiation values have been smoothed across color bands and that the general radiation trend is captured approximately along the second DM feature even if not every detail is modeled (recall that measured radiation is the target value and, thus, is not included in the DM transformation). This behavior can be observed for the training points in first and second plots but also for the test ones in the third and fourth ones, where it can be seen that the ALP method makes a good extension of the target function for new points.

For comparison purposes training and test ALP results when the DM embedding is computed over the entire sample are shown in Figure~\ref{embedding} (we could consider this as the ``true" DM embedding).
Comparing with Figure~\ref{embeddingLP} we can see that the two DM embeddings are very similar and that the target and prediction colors seem to be more or less the same. 
This shows that when we apply ALP to compute the DM coordinates of new test sample points we get an embedding quite close to the ideal one obtained jointly over the train and test patterns.

In order to give a quantitative measure of the quality of the ALP projections, we perform Euclidean $K$-means with $K=3$ over the three-dimensional embeddings of the test sample. We want to compare the clusters obtained over the ideal DM embedding computed with the entire sample and the ones obtained over the train DM embedding and then extended for the test coordinates. The resulting clusterings can be seen in figure~\ref{clusters}.
Notice that these clusters do not reflect radiation structure; instead more weight is apparently given to the first DM feature (that should have the largest feature values and, thus, a bigger influence when Euclidean distances are computed). Anyway, recall that this embedding was made with $15$ different variable types, from which just $5$ are radiation variables. Because of this, the cluster structure doesn't have to reflect just radiation's effect on the embedding, but the overall variable behavior. 

To obtained a concrete metric we will compare the cluster assignments of the test points over the extended embedding (the ``predicted" assignments) with those made over the full embedding (the ``true" assignments).
Looking at this as a classification problem, the accuracy, i.e., the percentage of test points assigned to their ``real" clusters, measures how well the ALP extensions match the ``true" embedding structure. 
As the confusion matrix in Table~\ref{CMatrix} shows, ALP does this quite well, with a total accuracy of $97.53\%$.
Thus, if clustering of the embedded features is desired, ALP makes it possible for new patterns with a very high accuracy. 
\begin{table}[ht]
\ra{1.4}
\ca{3}
\caption{Confusion Matrix of the $K$-means classification for the extended test coordinates.}
\begin{center}
\begin{tabular}{@{} l | c c c | c @{}}
& P. $1$ & P. $2$ & P. $3$ & $\sum$ \\
\toprule
R. $1$ & $294$ & $4$ & $0$ & $298$ \\
R. $2$ & $9$ & $342$ & $2$ & $353$ \\
R. $3$ & $0$ & $12$ & $432$ & $444$ \\
\cline{1-5}
$\sum$ & $303$ & $358$ & $434$ & $1095$ \\
\bottomrule
\end{tabular}
\label{CMatrix}
\end{center}
\end{table}

Getting back to radiation prediction over the test sample, once the embedding is obtained, we try to predict over it the total daily incoming solar energy using now a two step ALP procedure. 
To do so, DM are applied over the training sample to obtain the DM embedded features and then a first ALP model ${\cal ALP}_F$ is built to extend the DM features to the test sample and a second ALP ${\cal ALP}_R$ to build the ALP function approximation to the target radiation. Given a new NWP test pattern, we apply ${\cal ALP}_F$ to obtain its extended DM features and, next, ${\cal ALP}_R$ over them to obtain the final radiation prediction.

\begin{figure}[ht!]
\centering
	\includegraphics[width=0.35\textwidth]{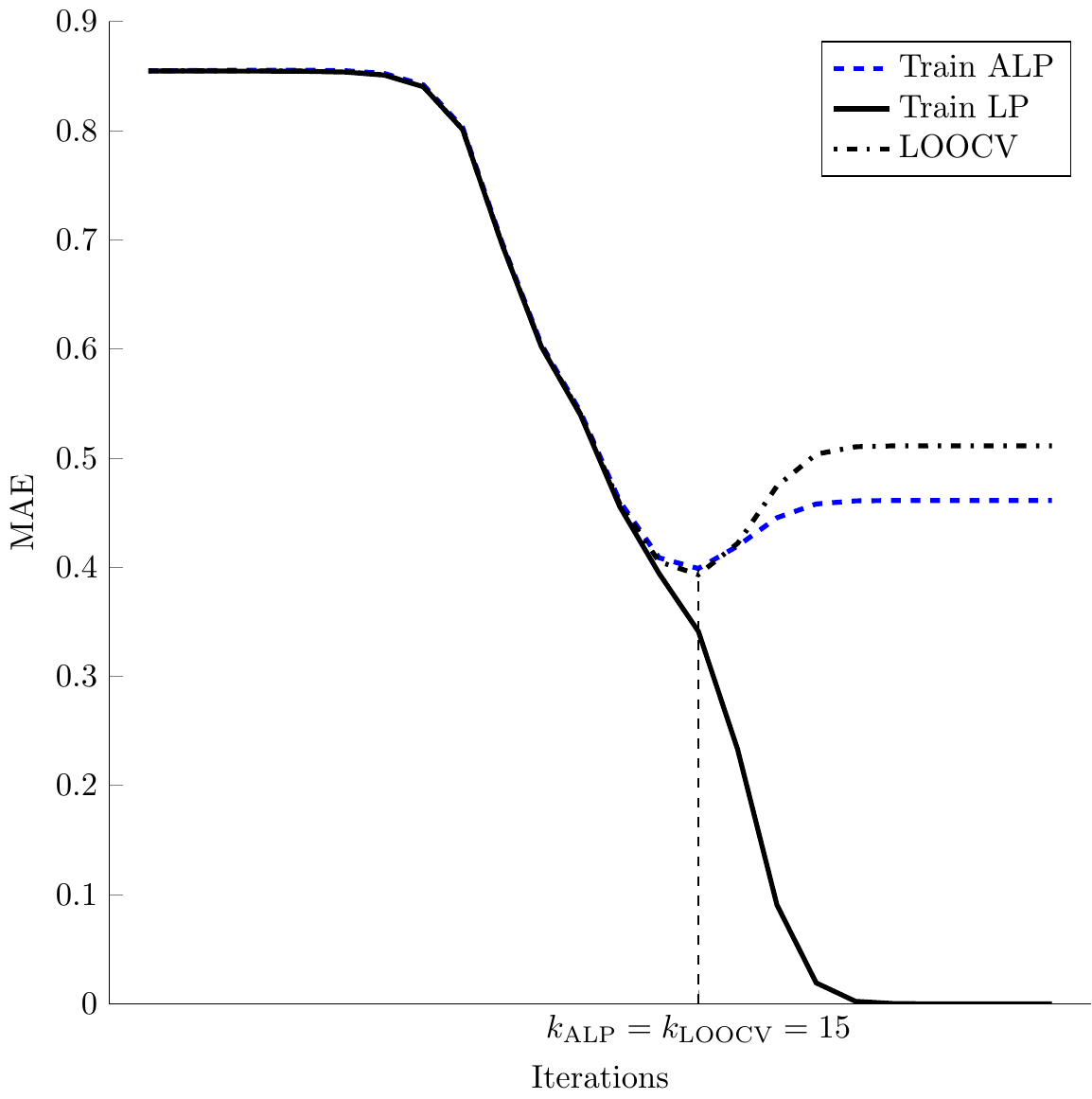}
\caption{Training error for the solar example of ALP, standard LP and its associated LOOCV.}
\label{errors} 
\end{figure}

\begin{figure}[ht!]
\centering
	\subfloat{\includegraphics[height=0.23\textwidth]{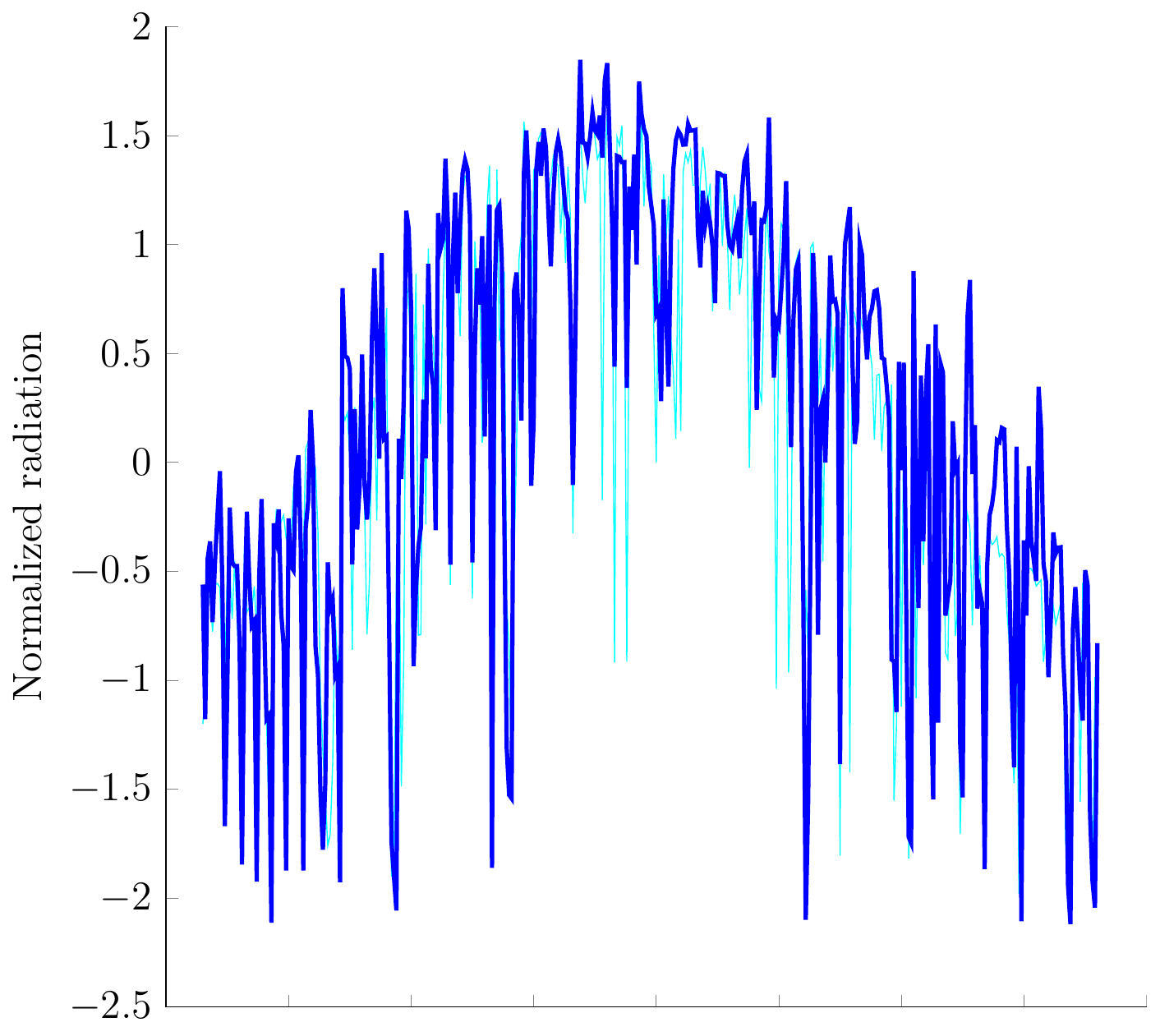}}
	\subfloat{\includegraphics[height=0.23\textwidth]{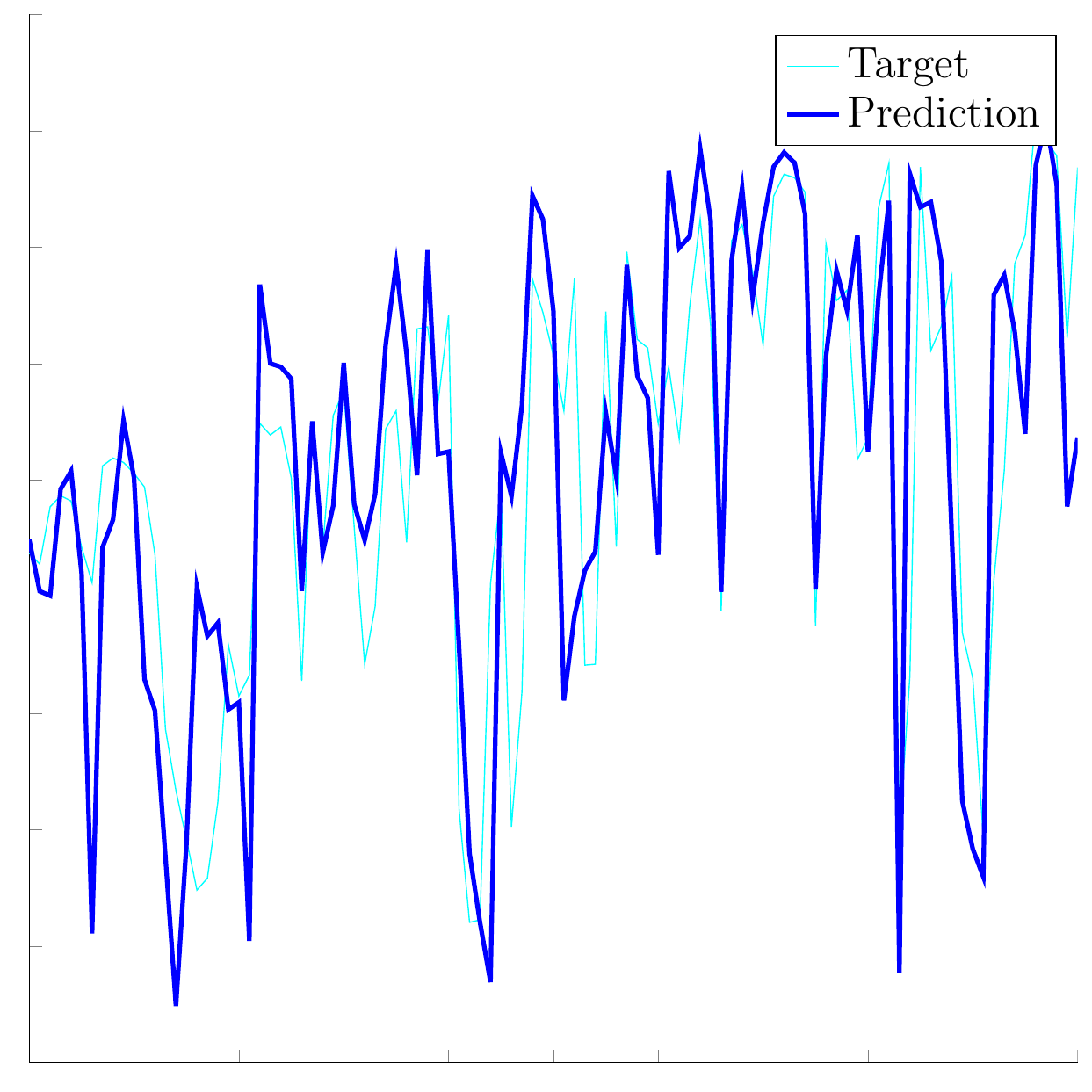}}
\caption{Prediction of the daily incoming solar energy over the second test year, and in a zoom over $100$ days.} 
\label{real} 
\end{figure}

In Figure~\ref{real}, left, we have depicted for the second test year the real radiation in light blue and the ALP prediction in dark blue. 
Although the winning models in the Kaggle competition followed different approaches, it can be seen that ALP captures radiation's seasonality. 
In the right plot we zoom in and it can be appreciated how ALP tracks the radiation variations. 
Even if not every peak is caught, ALP yields a reasonably good approximation to actual radiation without requiring any particular parameter choices nor any expert knowledge about the problem we wanted to address.
Figure~\ref{errors} shows the evolution of standard LP training error, its associated LOOCV error and the LOOCV estimation given by ALP when they are applied for radiation forecasting. It illustrates the robustness of the ALP model against overfitting. Again, the ALP model requires here the same number of $15$ iterations suggested by applying full LOOCV to standard LP.

\begin{figure*}[ht!]
\centering
	\subfloat{\includegraphics[width=0.33\textwidth]{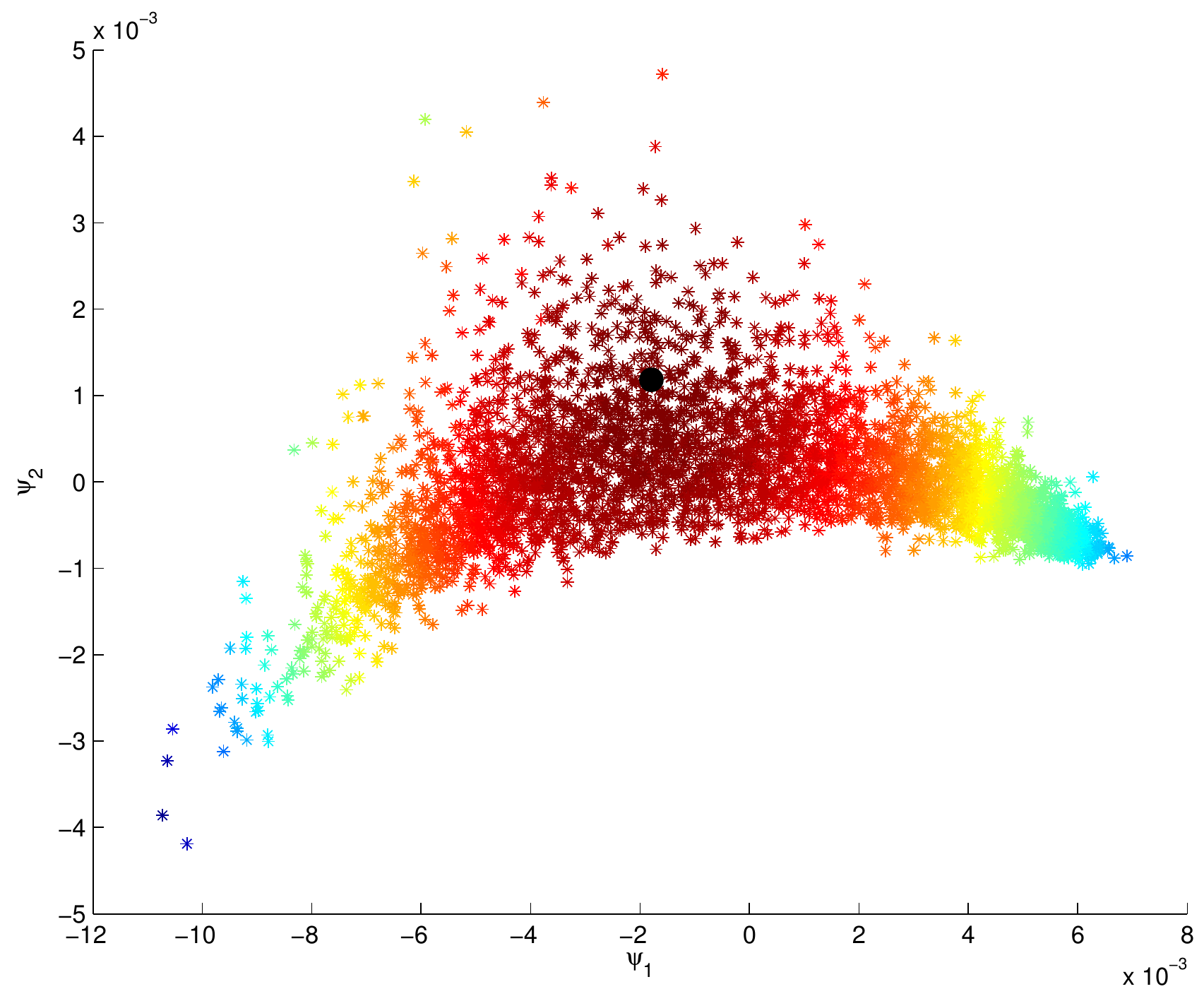}}
	\subfloat{\includegraphics[width=0.33\textwidth]{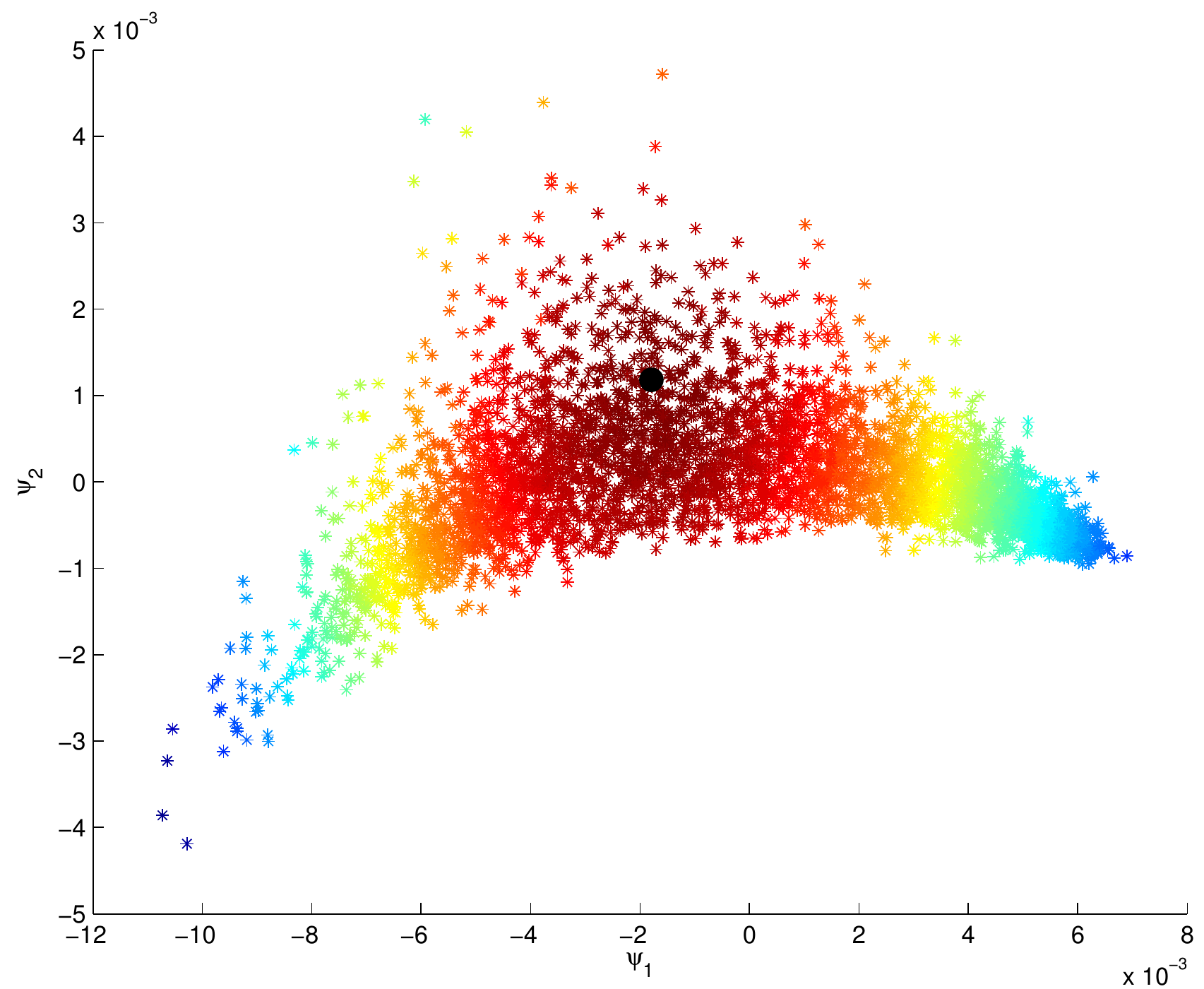}}
	\subfloat{\includegraphics[width=0.33\textwidth]{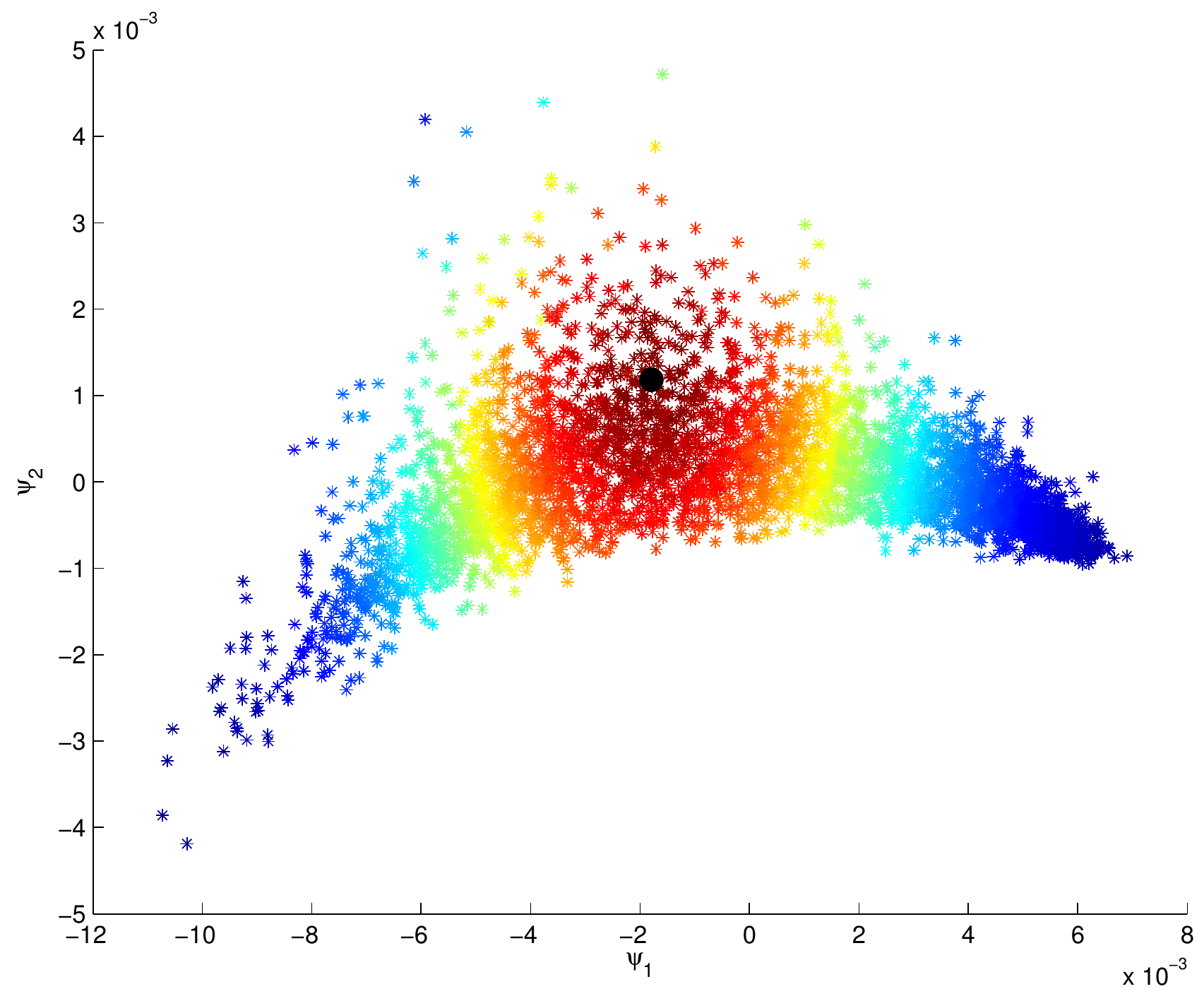}} \\
	\subfloat{\includegraphics[width=0.33\textwidth]{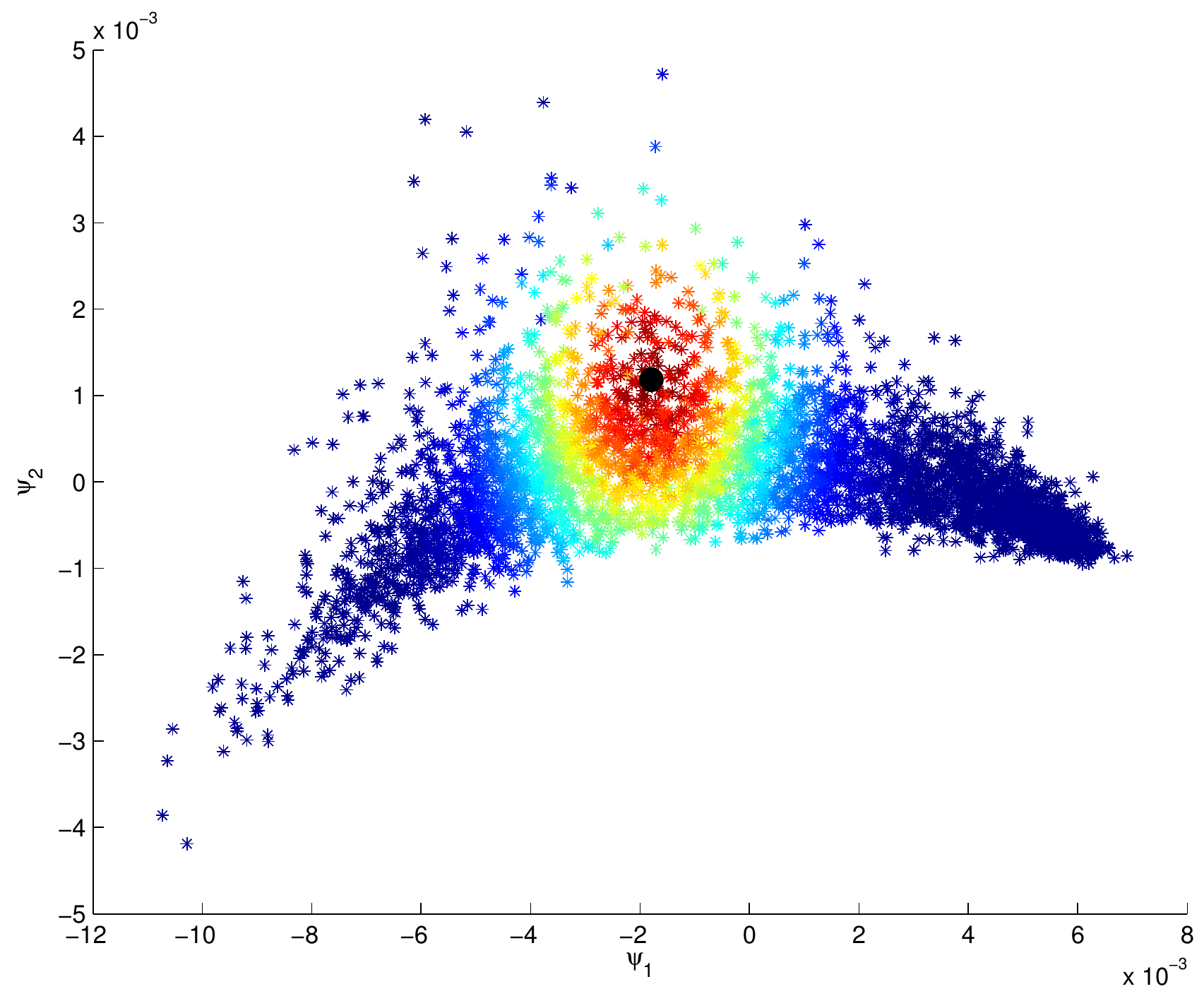}}
	\subfloat{\includegraphics[width=0.33\textwidth]{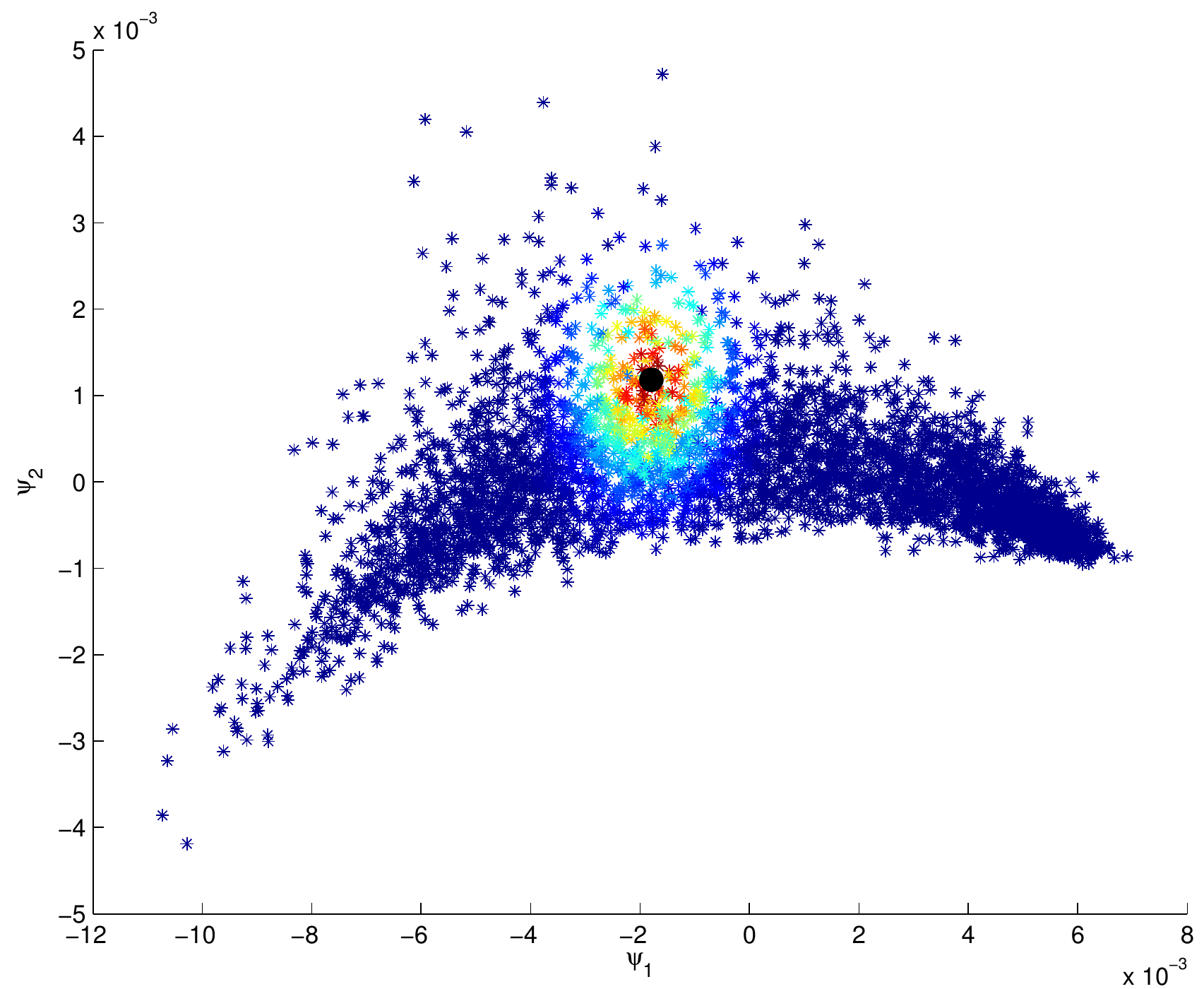}}
	\subfloat{\includegraphics[width=0.33\textwidth]{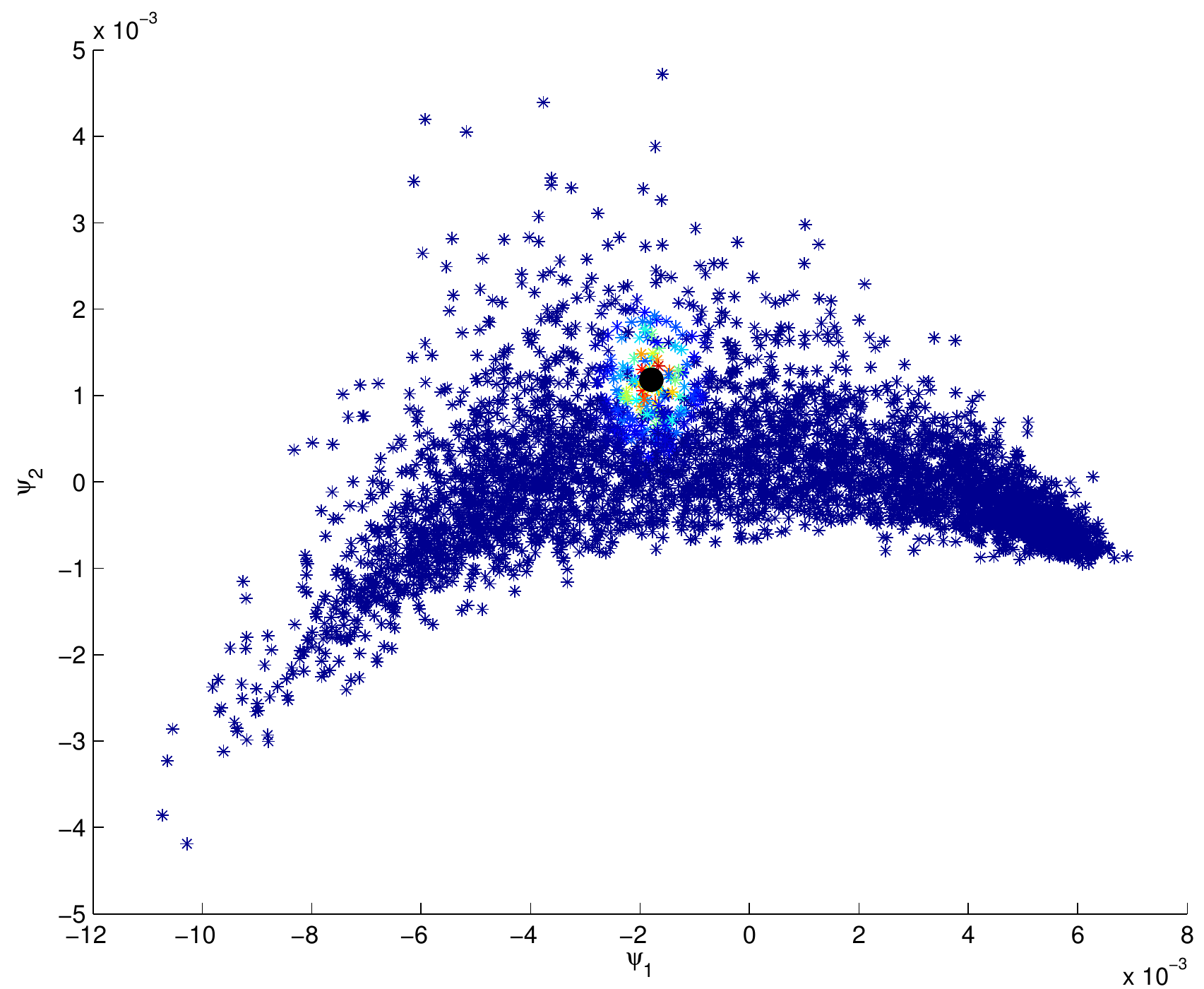}}
\caption{Evolution of the neighborhood of a test point over the embedded training points.} 
\label{sigma} 
\end{figure*}

Finally, Figure~\ref{sigma} shows for a test point $x$, plotted as a black dot, the evolution as ALP advances of the influence of sample points on $x$, larger for red points and smaller for blue ones. As it can be seen, as $\sigma$ gets smaller, the number of high influence red points also decreases sharply and so does the possibility of overfitting.

\section{Conclusions}
\label{sec:Concl}
The classical Laplacian Pyramid scheme of Burt and Adelson have been widely studied and applied to many problems, particularly in image processing. However, it has the risk of overfitting and, thus, requires the use of rather costly techniques such as cross validation to prevent it.

In this paper we have presented Auto-adaptative Laplacian Pyramids (ALP), a modified, adaptive version of LP training that yields at no extra cost an estimate of the LOOCV value at each iteration, allowing thus to automatically decide when to stop in order to avoid overfitting. 
We have illustrated the robustness of the ALP method over a synthetic example and shown on a real radiation problem how to use it to extend Diffusion Maps embeddings to new patterns and to provide simple, yet reasonably good radiation predictions.

\subsection*{Acknowledgments}
The authors wish to thank Prof. Yoel Shkolnisky and Prof. Ronald R. Coifman for helpful remarks. The first author acknowledge partial support from grant TIN2010-21575-C02-01 of Spain's Ministerio de Econom\'{i}a y Competitividad and the UAM--ADIC Chair for Machine Learning in Modeling and Prediction. 
\bibliographystyle{plain}
\bibliography{AutoadapLP}
\end{document}